\documentclass{article} 
\usepackage{iclr2024_conference,times}

\usepackage{amsmath,amsfonts,bm}

\def\eqref#1{equation~\ref{#1}}

\def\1{\bm{1}}

\DeclareMathAlphabet{\mathsfit}{\encodingdefault}{\sfdefault}{m}{sl}
\SetMathAlphabet{\mathsfit}{bold}{\encodingdefault}{\sfdefault}{bx}{n}

\usepackage{url}
\usepackage{multirow}
\usepackage{booktabs}
\usepackage{graphicx}
\usepackage{algorithm}
\usepackage{algorithmic}
\usepackage{listings}

\usepackage{wrapfig}

\definecolor{citecolor}{HTML}{2779af}
\definecolor{linkcolor}{HTML}{c0392b}
\usepackage[pagebackref=false,breaklinks=true,colorlinks,bookmarks=false,citecolor=citecolor,linkcolor=linkcolor]{hyperref}

\makeatletter

\def\blfootnote{\gdef\@thefnmark{}\@footnotetext}

\makeatother

\renewcommand\vec{\boldsymbol}

\newcommand\bx{\vec{x}}
\newcommand\ba{\vec{a}}
\newcommand\bt{\vec{t}}

\newcommand\bc{\vec{c}}

\newcommand\bh{\vec{h}}

\newcommand\bA{\vec{A}}

\newcommand\bmu{\vec{\mu}}
\newcommand\bsigma{\vec{\sigma}}
\newcommand\bepsilon{\vec{\epsilon}}

\newcommand\bs{\vec{s}}

\newcommand\bb{\vec{b}}

\title{Perceptual Group Tokenizer: \\ Building Perception with Iterative Grouping}

\author{Zhiwei Deng, Ting Chen, and Yang Li \\
Google Research and Deepmind \\
}

\iclrfinalcopy 
\begin{document}

\maketitle

\begin{abstract}

Human visual recognition system shows astonishing capability of compressing visual information into a set of tokens containing rich representations without label supervision. One critical driving principle behind it is perceptual grouping \citep{palmer2002perceptual, wagemans2012century, herzog2018perceptual}. Despite being widely used in computer vision in the early 2010s, it remains a mystery whether perceptual grouping can be leveraged to derive a neural visual recognition backbone that generates as powerful representations. In this paper, we propose \textit{the Perceptual Group Tokenizer}, a model that entirely relies on grouping operations to extract visual features and perform self-supervised representation learning, where a series of grouping operations are used to iteratively hypothesize the context for pixels or superpixels to refine feature representations. We show that the proposed model can achieve competitive performance compared to state-of-the-art vision architectures, and inherits desirable properties including \textit{adaptive computation without re-training}, and interpretability. Specifically, Perceptual Group Tokenizer achieves 80.3\% on ImageNet-1K \textit{self-supervised learning} benchmark with linear probe evaluation, establishing a new milestone for this paradigm.

\end{abstract}

\vspace{-4mm}
\section{Introduction}

Visual recognition mechanisms matter. The pursuit of advanced vision algorithms that encode an image to meaningful representations dates back to late 80s, with two paradigms marking the progress over the past 40 years: feature detection \citep{lecun1998gradient, lowe2004distinctive, he2016deep, liu2022convnet} and perceptual grouping \citep{shi2000normalized, uijlings2013selective, arbelaez2014multiscale}, where feature detection focuses on specific distinctive patterns, while perceptual grouping considers similarities among all pixels to produce a compact set of tokens as proxies for image representation. Ever since the surge of deep learning, feature detection has predominated the vision field and become the main principle behind
representation learning backbone designs and made impressive progress \citep{simonyan2014very, szegedy2015going, he2016deep, chen2017deeplab, tan2019efficientnet, qi2020deep, liu2022convnet}. The success of the former paradigm is, although striking, raising the question of whether perceptual grouping can also be used as the driving principle to construct a visual recognition model.

Different from detecting and selecting distinctive features, perceptual grouping emphasizes on learning feature space where similarity of all pixels can be effectively measured \citep{uijlings2013selective, arbelaez2014multiscale}. With such a feature space, semantically meaningful objects and regions can be easily discovered with a simple grouping algorithm and used as a compact set to represent an image \citep{uijlings2013selective, arbelaez2014multiscale, locatello2020object}. This indicates that image understanding is essentially ``pixel space tokenization'', and being able to produce generalizable feature representations is tightly connected to whether the correct contextual pixels are binded together \citep{hinton2022represent, culp2022testing}.

The intriguing properties of perceptual grouping, including natural object discovery, deep connections with information theory and compression \citep{ma2007segmentation}, and association with biological vision system \citep{herzog2018perceptual} or cognitive science explanations \citep{palmer2002perceptual}, have led to a strong revival recently under deep learning frameworks \citep{locatello2020object, elsayed2022savi++, xu2022groupvit, wu2022slotformer, biza2023invariant}. However, these methods are either still focusing on small or toy datasets \citep{locatello2020object, chang2022object, biza2023invariant}, or used as an auxilliary component \citep{xu2022groupvit, ke2022cast, seitzer2022bridging} to strengthen existing vision architectures for increased interpretability. Whether perceptual grouping can be used to build models and learn representations that are as informative and expressive as those learned by state-of-the-art vision architectures remains an open question.

In this paper, we propose \textit{Perceptual Group Tokenizer}, a model trained under a \textit{self-supervised learning} framework, which builds visual representation \textit{entirely based on perceptual grouping operations}. Given an image, the core of our model is to understand each pixel or patch through hypothesizing its contexts with grouping operations. Starting from given input patches, the grouping operation performs an iterative binding process onto a set of randomly sampled group tokens to determine the affinity groups based on similarities. The group tokens are then used as hypothesized contexts to refine the feature representation for the image. We show that applying this simple principle can already produce expressive representations and works well with self-supervised pretraining on a large vision dataset.

The grouping operation is also closely related to self-attention, a highly popular method commonly used in modern vision backbones. We build connection between the proposed grouping operation and self-attention and show that, if group tokens are treated as communication channels, self attention can potentially automatically emerge during learning processes as a special case, while the grouping operation can produce even richer interactions among tokens. Under this viewpoint, ViT \citep{dosovitskiy2020image} can be considered as a grouping backbone, with a fixed number of grouping slots equal to the number of input tokens, and the binding is achieved through stacking more than one layer with non-shared weights. This provides one explanation on why grouping mechanism can be effective on visual representation learning and has the potential to be a promising competitive paradigm for vision architecture designs.

The primary contribution of this work is proposing a new architecture derived purely by perceptual grouping that achieves competitive performance compared to other state-of-the-art architectures on \textit{self-supervised learning} benchmarks, contributing to a new paradigm of developing vision architectures. The model has several key differences and advantages over ViT, including (1) explicit separating out the ``group token'' concept to allow for automatic image parsing and flexible customization on the number of groups without being binded to the number of patches; (2) much less peak memory usage during inference time given the same number of input tokens; (3) adaptive computation without re-training the model, leading to flexible usage according to domains and computes.

\begin{figure}[t]
    \centering
    \includegraphics[width=1.0\linewidth]{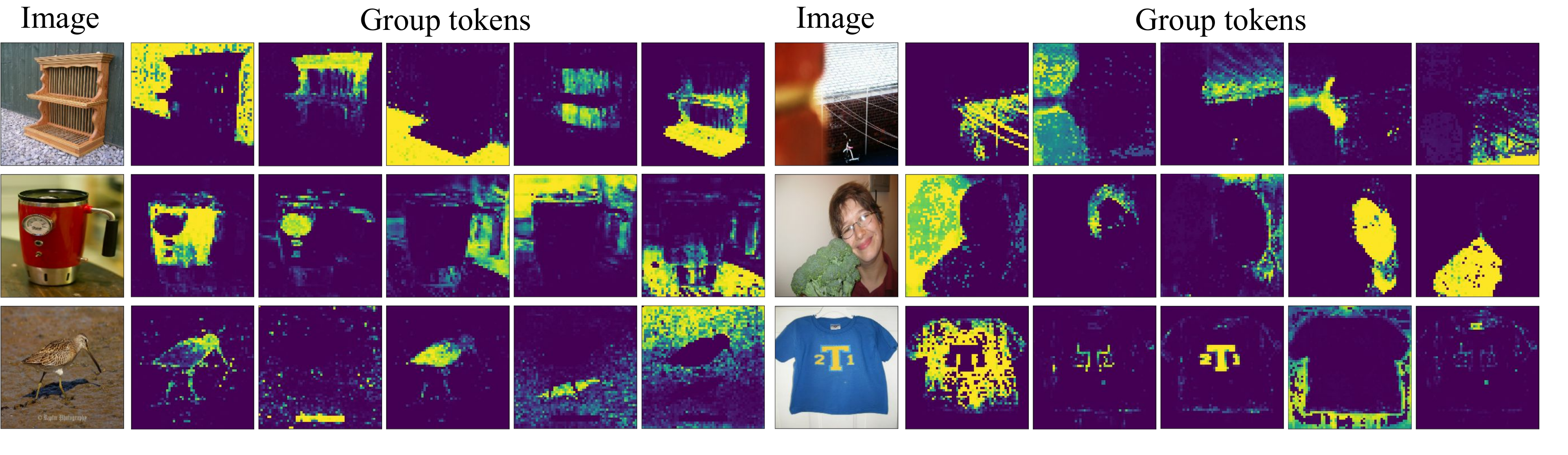} 
    \vspace{-6mm}
    \caption{\textbf{Perceptual Group Tokenizer} is entirely driven by \textbf{grouping operations} to perform representation learning. Group tokens (discovered objects) are shown above. See more in the appendix.}
    \label{fig:pull_fig}
    \vspace{-4mm}
\end{figure}

\vspace{-2mm}
\section{Related works}
\vspace{-2mm}
\textbf{Vision architectures.} There are two main frameworks for vision backbones. The first framework is Convolutioinal neural networks, which rely on local filters, sliding windows and translational equivariance to perform representation learning. Since the introduction of ConvNets in 1980s, ConvNet was repopularized by AlexNet \citep{krizhevsky2012imagenet}. The line of ConvNet is a classical inheritance from traditional feature detection methods \citep{lowe2004distinctive, dalal2005histograms, rosten2008faster}, where instead of hand crafting features, an overcomplete set of filters are automatically learned to obtain high-response regions. The object understanding is built along the depth axis \citep{simonyan2014very, szegedy2015going, he2016deep}, with early layers capturing low-level parts and higher-level layers producing object structure representations \citep{zeiler2014visualizing, zhou2014object, yosinski2015understanding, bau2017network}.  In the feature detection framework, not every pixel is worth being used depending on particular tasks, leading to difficulty in obtaining representation for each pixel. 

Recently, Vision Transformer (ViT) \citep{dosovitskiy2020image}, a second vision backbone framework, shows impressive performance and has surpassed ConvNet on visual recognition. The core of ViT is the iterative applying of self-attention operations \citep{vaswani2017attention, dosovitskiy2020image}. A direct usage of ViT on small patches (thus a high-resolution grid) is extremely computationally expensive due to its associated quadratic cost. Therefore, a common practice is often partitioning the image into large non-overlapping patches \citep{dosovitskiy2020image, touvron2021training}, or constrain the operation to local regions \citep{liu2021swin}. 

\textbf{Self-supervised learning.} The field of representation learning has seen significant interest in self-supervised learning during the past few years. The main evaluation results using linear probe on ImageNet benchmarks is approaching the results obtained by supervised learning  \citep{oquab2023dinov2}. Contrastive representation learning is the early method that shows promising results \citep{oord2018representation, chen2020simple, tian2020contrastive}. BYOL \citep{grill2020bootstrap} and DINO \citep{caron2021emerging} propose to use a moving average target of an online network to perform self representation matching. Masked image modeling also shows to be effective on representation learning, where the masking is either at the pixel level \citep{he2022masked} or the learned codebook level \citep{bao2021beit}. 

\textbf{Object discovery.} The perceptual grouping is essentially performing ``object and stuff'' discovery in the pixel space. It has broad connections with the early works in computer vision \citep{shi2000normalized, uijlings2013selective, levinshtein2013multiscale,  arbelaez2014multiscale, pont2016multiscale}, the recent progress on object-centric representation \citep{burgess2019monet, locatello2020object, chang2022object, hinton2022represent, henaff2022object, culp2022testing, elsayed2022savi++}, and biological or neural mechanisms on perceptual grouping \citep{palmer2002perceptual, wagemans2012century, herzog2018perceptual, kim2019disentangling}. Despite the early popularity of perceptual grouping methods on various computer vision tasks \citep{shi2000normalized, uijlings2013selective, levinshtein2013multiscale, krahenbuhl2011efficient}, it has not attracted significant attention until several recent works that apply it as a side component on top of another main backbone \citep{seitzer2022bridging, liu2022dynamic, xu2022groupvit, ke2022cast}. Some relevant works demonstrate alternative possibilities in architecture design, but only uses cross attention without refining the patch feature space \citep{jaegle2021perceiver}, 
or apply it on diffusion tasks \citep{jabri2022scalable}. Other methods also attempt to use ad-hoc sparsification methods on top of ViT \citep{rao2021dynamicvit, yin2022vit, bolya2023token} for efficiency and are orthogonal to our work. A most related work \citep{ma2023image} focuses on supervised learning and relies on fixed-center pooling and less standard operations. In our proposed model, we adopt a design as ViT except for self attention, and highlight several key technical contributions, including multi-grouping with multi-seeding, adaptive computation without re-training, and other design choices for self-supervised representation learning.



\vspace{-2mm}
\section{Models}
\vspace{-3mm}
In this section, we introduce Perceptual Group Tokenizer (PGT), a visual recognition architecture entirely driven by perceptual grouping principles. We discuss the core operations for grouping in section \ref{sec:grouping}, the building blocks and network architectures in section \ref{sec:arch}, the loss function used for self-supervised learning in section \ref{sec:loss}, and the connections with other models in section \ref{sec:discuss}.

\vspace{-3mm}
\subsection{Perceptual grouping}
\label{sec:grouping}


\begin{figure}[t]
    \centering
    \includegraphics[width=1.0\linewidth]{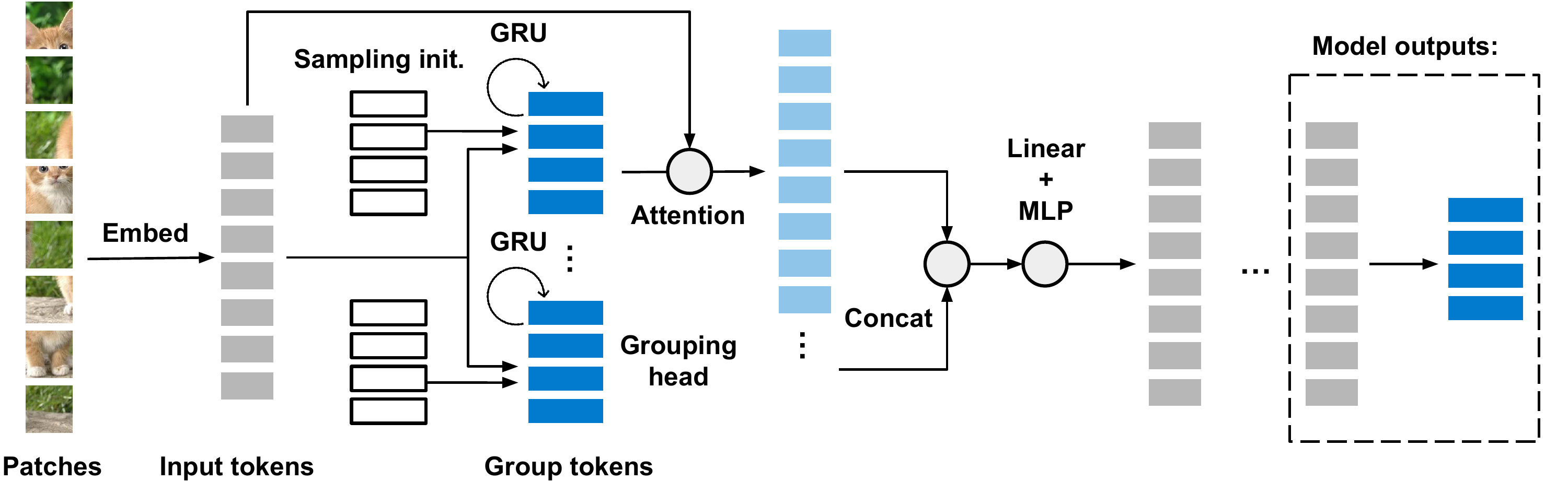}
    \vspace{-8mm}
    \caption{Perceptual Group Tokenizer takes in a sequence of patches (or pixels), generates high-dimensional embedding vectors for all patches, then them passes through a series of grouping layers to refine the embedding vectors as feature representations. Each grouping layer performs $K$ rounds of binding from input tokens to group tokens. To consider various grouping possibilities, multiple grouping heads are adopted. Each group token provides a useful context for input tokens for feature refinement. The final output of the model contains refined input token, group tokens, and assignments between input tokens and groups tokens.}
    \label{fig:main}
    \vspace{-6mm}
\end{figure}

We start with introducing notations for our method. Given an image $\bx \in \mathbb{R}^{H\times W\times C}$, we first reshape it as a sequence of small patches\footnote{We use 4$\times$4 patches as inputs in this work. Note that our method is generalizable to either pure pixels or other forms of superpixels given a proper patch-to-vector embedding layer.}. Each patch $\bx_p \in \mathbb{R}^{h\times w\times c}$ has spatial shape $h\times w$, where $h\ll H$ and $w \ll W$, leading to $N=\frac{HW}{hw}$ number of patches per image. To represent a patch, we embed it into a high-dimensional vector $\bh \in \mathbb{R}^d$. The set of embedded tokens $\{\bh_i\}^N$ is referred to as \textit{input tokens} in later parts, and used as inputs for the following grouping blocks.



\textbf{Feature refinement through hypothesizing contexts.} Individual pixels do not have meanings without putting it into contexts. At a high level, image understanding or feature learning is equivalent to binding the correct contextual pixels at all locations. The core idea of our model is to generate many (e.g. over-complete w.r.t number of objects in the image) hypothesized contexts and use the hypothesized contexts as cues to refine the feature representation of each patch. This process is achieved through a grouping module. Given input tokens $\{\bh_i\}^N$, the grouping module starts from a set of random samples (referred as \textit{group tokens}) from a random distribution, then performs binding process to aggregate information from input tokens to the group tokens, and ends up with a set of group tokens $\bc^*=\{\bc_j^*\}_{j=1}^M$ representing hypothesized contexts among input tokens. The relation between $\bh_i$ and $\bc_j$ is soft assigment, indicating how likely an input token belongs to that context. Note that there are often \textit{various ways of generating groupings for an image}, e.g. different semantics, colors, textures, etc., we propose the ``multi-grouping operation'' to hypothesize rich contexts for tokens. The overall model is shown in figure \ref{fig:main}.


\textbf{Multi-grouping operation.} The building block of our model is the multi-grouping operation $\mathcal{G}$, which contains multiple heads to perform the binding process in parallel. This design encourages the model to consider multiple ways of generating groups under different projection spaces. Each head owns a separate Gaussian distribution with learnable means and variance, similar to \citep{kingma2013auto, locatello2020object}. Starting from a set of randomly sampled initial group tokens 
$\bc^{(0)}_{\small{\textsc{head}}} \sim p_{\textsc{init}}(\cdot)$,
the grouping operation uses doubly normalized attention weights to aggregate information from $\bh$, and the produced group tokens $\bc^{(1)}_{\small{\textsc{head}}}$ are used for the next round binding. The attention normalization and feature projection are performed in all heads separately.
\begin{eqnarray}
\label{eqn:grouping}
    \bc^{(1)}_{\text{\textsc{head}}} &=& \mathcal{G} (\bc^{(0)}_{\text{\textsc{head}}},\bh; \theta) \\
    &\cdots& \notag\\
    \bc^{*}_{\text{\textsc{head}}} = \bc^{(K)}_{\text{\textsc{head}}} &=& \mathcal{G} (\bc^{(K-1)}_{\text{\textsc{head}}},\bh; \theta)
\end{eqnarray}
where after K steps the final group tokens $\bc^{*}=\bc^{(K)}$ is obtained, and $\theta$ is learnable parameters in $\mathcal{G}$. The grouping operator is summarized in algorithm \ref{alg:grouping}. 




The sampling distribution $p_{\textsc{init}}(\cdot)$ for initializing group tokens $\bc^{(0)}_{\text{\textsc{head}}}$ needs to be lightweight. We explore two variations: (1) Gaussian distribution $p(\bmu_{\small{\textsc{head}}}, \bsigma_{\small{\textsc{head}}})$ with learnable means and variance, and a one-step normalizing flow module that transforms a unit Gaussian noise to a sample that follows more complex distributions. More details can be found in the appendix in section \ref{sec:init_dist}

\textbf{Implicit differentiation.} The iterative grouping process unrolls $K$ steps per operation and leads to heavy burden in the training computation graph. Instead of explicitly backpropagating through the unrolled graph, we follow \citep{chang2022object} and treat the multi-grouping process as a fixed point iteration per head. The gradient in the backpropagation is approximated using first-order Neumann series, which can be simply achieved by detaching the output before the final iteration. 

\begin{algorithm}[tb]
\caption{Multi-grouping operation using $\mathcal{G}$.
}
\label{alg:grouping}
\definecolor{codeblue}{rgb}{0.25,0.5,0.5}
\definecolor{codekw}{rgb}{0.85, 0.18, 0.50}
\lstset{
  backgroundcolor=\color{white},
  basicstyle=\fontsize{7.2pt}{7.2pt}\ttfamily\selectfont,
  columns=fullflexible,
  breaklines=true,
  captionpos=b,
  commentstyle=\fontsize{7.2pt}{7.2pt}\color{codeblue},
  keywordstyle=\fontsize{7.2pt}{7.2pt}\color{codekw},
  escapechar={|}, 
}
\vspace{-1mm}
\begin{lstlisting}[language=python]
def multi_grouping(h_key, h_value, steps, num_tokens, num_heads):
  """ Input tensors:
         h_key and h_value are projected multi-head tensors with shape [num_heads x N x d].
  """
  # Initial M group tokens.
  group_tokens = sampling_distribution(nsamples=num_tokens, choice='Gaussian') # or 'Flow'
  group_tokens = group_tokens.reshape(num_heads, num_tokens, d) #[num_heads x M x d]

  # Binding process.
  for step in range(steps):
    # Implicit differentiation.
    if step == steps - 1:
      group_tokens = stop_gradient(group_tokens)
    """ The following is a one-step grouping operation. """
    # Attention operation for group assignment.
    attn_matrix = attention(group_tokens, h_key) #[num_heads x N x M]
    attn_matrix /= attn_matrix.sum(-2, keep_dim=True)
    h_updates = einsum("hij,hid->hjd", attn_matrix, h_value) #[num_heads x M x d]
    group_tokens = gru_cell(h_updates, group_tokens)
    # Grouped mlp/layernorm performs independent mlp/layernorm for each head.
    group_tokens = grouped_mlp(grouped_layer_norm(group_tokens)) + group_tokens
    
  return group_tokens
\end{lstlisting}
\vspace{-2mm}
\end{algorithm}



\subsection{Network architecture}
\label{sec:arch}

Similar to standard ViT, our model refines the hidden representation $\bh$ using $L$ model layers. We use $\bh^{l}$ to denote the representation after each layer, and explain the design in this section.

\textbf{Grouping layer.} Each grouping layer takes in $\bh^{l-1}$ as input, and uses the grouping operation in equation \ref{eqn:grouping} to generate group tokens $\bc_{\textsc{head}}^*=\{\bc_{j,\textsc{head}}^*\}_{j=1}^M$. To use the group tokens to provide context for each $\bh_i^{l-1}$, we perform another attention operation to obtain the attention matrix (only normalized over group token axis) $\bA \in \mathbb{R}^{N\times M}$ representing the assignment from input tokens to group tokens, and aggregate the feature back to the input token space:
\begin{eqnarray}
    \bh_{\textsc{head}}^{l} &=& \bA [\bc_{1,\textsc{head}}^*; \bc_{2,\textsc{head}}^*; ...; \bc_{M,\textsc{head}}^*] \\
    \bh^{l} &=& \text{Linear}([\bh_{\textsc{head}_1}^{l};...\bh_{\textsc{head}_H}^{l}]) \\
    \bh^{l} &=& \bh^{l-1} + \text{MLP}(\text{LN}(\bh^{l}))
\end{eqnarray}
This layer definition follows the standard ViT layer as close as possible, where features from each head are aggregated through concatenation and a linear layer transformation. Each token $\bh$ is further refined using a follow up multi-layer perceptron.

\textbf{Grouping blocks.} Similar to previous architecture designs \citep{he2016deep, liu2021swin}. we define blocks for the model. One block contains multiple grouping layers that share the same hyperparameters setups, i.e. the number of group tokens, and group token dimensions. The full model contains three grouping blocks. This increases the flexibility when exploring model design spaces.



\vspace{-3mm}
\subsection{Self-supervision loss}
\label{sec:loss}

We strictly follow the student-teacher self-supervision loss \citep{caron2021emerging, oquab2023dinov2}, and use a moving average of online network (student model) as the teacher model to perform representation learning. To summarize group tokens outputed from the final layer, we use one multi-head attention layer with a learnable token to attend to all group tokens. The produced single vector is treated as the feature representation for the image and is input to the loss function.


\begin{wrapfigure}{r}{0.38\textwidth}
\vspace{-5mm}
  \begin{center}
    \includegraphics[width=0.4\textwidth]{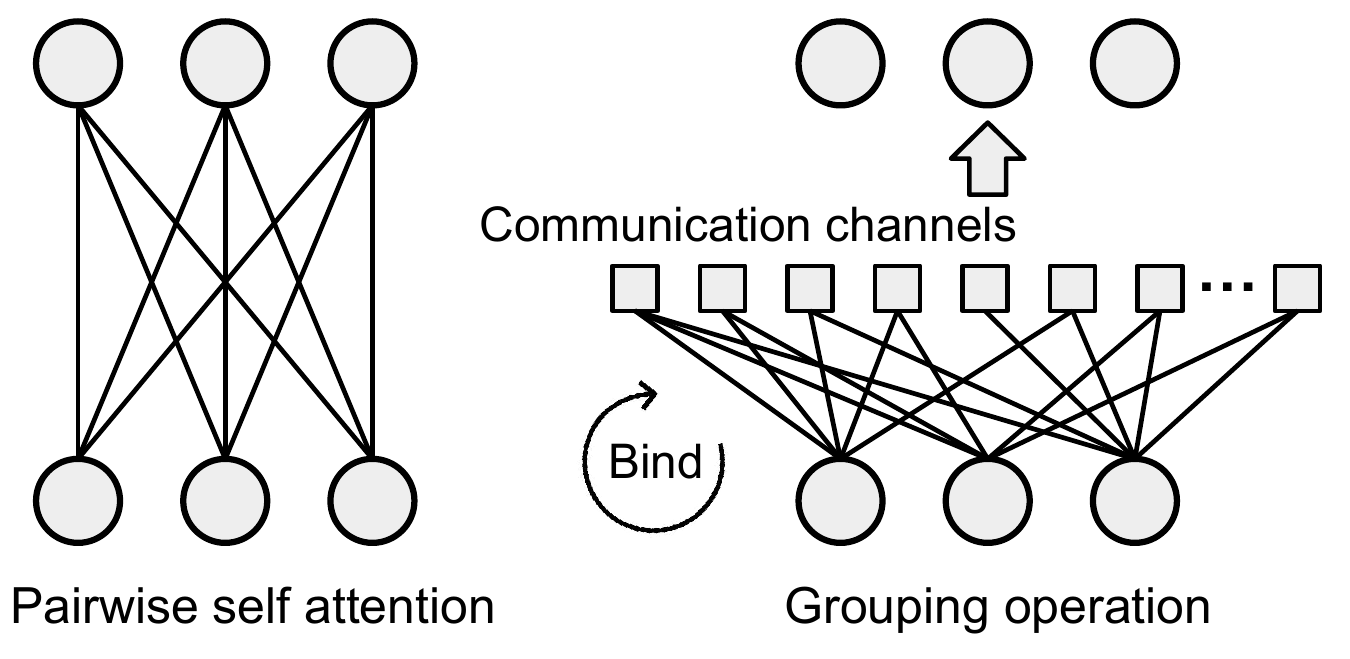}
  \end{center}
  \vspace{-6mm}
\caption{Operation comparison.}
\label{fig:communication_channel}
\vspace{-5mm}
\end{wrapfigure}

\vspace{-3mm}
\subsection{Discussion}
\label{sec:discuss}
Our proposed model, perceptual group tokenizer, does not contain self-attention operations and purely relies on grouping operations. In this section, we link the grouping process to several techniques and discuss the rationale on why this model can be effective on representation learning.

\textbf{Group tokens as ``communication channels''.} The core of feature representation learning is how information is exchanged among pixels. In perceptual grouping backbones, we can consider the set of group tokens as communication channels, where information from different input tokens are aggregated in various ways. Each group token represents a high-order channel that links input tokens with high affinity under certain projected space to exchange information among them. As a thought experiment, if each input token is solely assigned to a different group token (given enough group tokens), then the perceptual grouping layer is equivalent to one self attention layer (up to some engineering design difference). While self attention layers mainly rely on pairwise communications, grouping operation, hypothetically, can automatically learn and emerge both pairwise and higher-order information exchange through the group token communication channels. This can also be linked to traditional \textit{factor graphs} in probabilistic graphical models. Through the lens of that, grouping is forming factor nodes automatically through the learning processes. With a properly designed loss and grouping operation, it has the potential to be more effective if adopting a per-layer comparison with self-attention operations.






\textbf{Efficiency.} Due to the flexibility in customizing number of group tokens (controlled by initial number of samples), grouping operation does not require a strict $O(N^2)$ operation and is $O(NM)$ on complexity. Furthermore, we show that \textit{ in inference time}, number of group tokens can \textit{even be adaptively customized}, given an already trained model.

\vspace{-3mm}
\section{Experiments}
\vspace{-3mm}
We evaluate the representation learned by our model on standard benchmarks based on the ImageNet-1K dataset. We also explore and analyze the design space of perceptual group tokenizer in section \ref{sec:ablations}, investigate its adaptive computation ability in section \ref{sec:adapt}, demonstrate its generalization ability on semantic segmentation in section \ref{sec:seg}, and visualize learned attentions in section \ref{sec:grouping_vis}.









\begin{table}[!b]
  \begin{center}
    \begin{tabular}{ccccc}
    \toprule
      Method & Arch & Param. & Linear probe (top-1 acc)\\
      \midrule
      \multicolumn{4} {l} {(\textit{Other backbones with different losses within the same batch of DINO for reference})} \\
      SCLR \citep{chen2020simple} & RN50W4 & 375 & 76.8 \\
      SwAV \citep{caron2020unsupervised} & RN50W2 & 93 & 77.3 \\
      BYOL \citep{caron2020unsupervised} & RN50W2 & 93 & 77.4 \\
      SwAV \citep{caron2020unsupervised} & RN50W5 & 586 & 78.5 \\
      BYOL \citep{caron2020unsupervised} & RN50W4 & 375 & 78.6 \\
      iBOT \citep{zhou2021ibot} & ViT-B/16 & 85 & 79.5 \\
      BYOL \citep{caron2020unsupervised} & RN200W2 & 250 & 79.6 \\
      SCLRv2 \citep{chen2020big} & RN152w3+SK & 794 & 79.8 \\
      BEiTv2 \citep{peng2022beit} & ViT-B/16 & 85 & 80.1 \\
      \midrule
      \multicolumn{4} {l} {(\textit{Fair comparison under the DINO loss and framework})} \\
      DINO \citep{caron2021emerging} & ViT-S/8 & 21 & 79.7 \\
      Ours (PGT$_{\textsc{G}}$-S-1024) & PGT-S & 34 & 79.8 \\
      DINO \citep{caron2021emerging} & ViT-B/16 & 85 & 78.2 \\
      DINO \citep{caron2021emerging} & ViT-B/8 & 85 & 80.1 \\
      Ours (PGT$_{\textsc{G}}$-B-256) & PGT-B & 70 & 79.7 \\
      Ours (PGT$_{\textsc{G}}$-B-512) & PGT-B & 70 & 79.9 \\
      Ours (PGT$_{\textsc{G}}$-B-1024) & PGT-B & 70 & 80.1 \\
      Ours (PGT$_{\textsc{F}}$-B-256) & PGT-B & 115 & 80.0 \\
      Ours (PGT$_{\textsc{F}}$-B-512) & PGT-B & 115 & 80.1 \\
      Ours (PGT$_{\textsc{F}}$-B-1024) & PGT-B & 115 & \textbf{80.3} \\
      \bottomrule
    \end{tabular}
    \caption{Comparison with strong baselines on ImageNet-1K under linear probe evaluation protocal. PGT$_{\textsc{Dist}}$-B-$X$ represents $X$ number of group tokens per grouping layer in inference (same trained model with 256 tokens is used). $\textsc{Dist}$: the distribution choice for group token initialization. $\textsc{G}$ and $\textsc{F}$ represent Gaussian and Flow, respectively. Our model achieves 80.3\%, competitive with state-of-the-art vision backbones.}
    \label{tab:main}
  \end{center}
  \vspace{-5mm}
\end{table}

\vspace{-3mm}
\subsection{Main results}
\vspace{-3mm}

\textbf{Setup.} The widely-adopted standard benchmark for evaluating self-supervised learning methods is ImageNet ILSVRC-2012 (ImageNet-1K) \citep{russakovsky2015imagenet}. Performance of models are measured by top-1 classification accuracy. The pre-trained backbones are frozen, with a linear classifier trained on top. For fair comparison, we follow the standard data augmentation used in \citep{caron2021emerging}, with the same number of global views and local views. The model is optimized using AdamW \citep{loshchilov2018fixing} with learning rate 0.0005 and 1024 batch size for 600 epochs, trained with TPUv5 for 21k core hrs (512 cores for 41 hrs). We use 4$\times$4 patches as image tokens, which keeps as much details as possible while maintaining reasonable computation costs.

\textbf{Architecture details.} In the experiments, we mainly evaluate two variants of PGT: the main model and a tiny version for exploring design choices. On the ImageNet-1K benchmark, we report the performance metrics of our main model. Three grouping blocks are used, with 10 grouping layers in each block. The dimension for input token is 384, with 256 group tokens per layer. The dimensions for group tokens are 98, 192, and 288 for the three blocks, respectively. There are 6 grouping heads used. For number of grouping iterations, we observe three rounds are sufficient to achieve good performance. The MLP hidden size for each layer is 384 as well, i.e., the MLP multiplication factor is 1. The final multihead attention layer uses a learnable token with 2048 dimensions to summarize all group tokens outputs from the model.

The main results are summarized in table \ref{tab:main}. We mainly compare with ResNet and ViT backbones, the two main stream vision architectures to show that perceptual grouping architecture can also achieve competitive results on the challenging ImageNet-1K benchmark. Although our model is trained with 256 group tokens, the model can use different numbers of group tokens in inference (more experiments in section \ref{sec:ablations}). We evalaute PGT with 256, 512, and 1024 number of group tokens and observe that the model can achieve 80.3\% top-1 accuracy, showing the self-supervised learned feature of PGT is as good as the ones learned by ViT architectures.
\vspace{-4mm}

\subsection{Ablations}
\vspace{-3mm}
To explore design choices of PGT, we use a tiny version of PGT with 3 blocks, 2 layer in each block (6 layers in total), 256 hidden size for input tokens, and 3 number of grouping iterations. The learnable token in MAP head has 512 dimensions. There are $\sim$10M parameters in this PGT-tiny.

\label{sec:ablations}

\textbf{Group token layouts.} Given a fixed number of budget on group tokens, we explore three choices on how they should be arranged across grouping blocks and layers: descend, flat and ascend. Intuitively, more group tokens will have higher capacity of capturing smaller parts and detailed visual features, while less group tokens are more prone to carry global information. As shown in table \ref{tab:design} bottom row, flat or descend number of group tokens performs the best. In practice, we find that using flat (same number of group tokens in three grouping blocks) version achieves better training stability.

\textbf{Group token dimension shapes.} Similar to token number arrangements, we explore how group token dimensions should be set. Under three choices, progressively increasing the dimension size in the later layers performs the best, shown in first row of table \ref{tab:design}. This also aligns with the intuition that later layers contain more information and requires higher capacity to represent groups.

\textbf{Multi-grouping vs single grouping.}  We further test whether multi-head grouping helps improve performance. As a fair comparison, we use 6 heads and 128 group tokens per head for a multi-grouping model, and 1 head with 6$\times$128 group tokens for a single grouping model. We find that adopting multi-head design can improve the performance from 62.2\% to 66.3\%, a 4.1\% accuracy boosts, showing that having multiple heads indeed helps with representation learning.

\begin{table}
  \begin{center}
    \begin{tabular}[b]{cccc}
        \toprule
         & Descend & Flat & Ascend \\
        \midrule
        Token size & \big[576, 384, 192\big] & \big[384, 384, 384\big] & \big[192, 384, 576\big] \\
        Accuracy & 62.0 & 63.1 & \textbf{63.4} \\
        \midrule
        Token shape & \big[192, 128, 64\big] & \big[128, 128, 128\big] & \big[64, 128, 192\big] \\ 
        Accuracy & 63.6 & \textbf{63.7} & 63.1 \\
        \bottomrule
    \end{tabular}
    \caption{Exploring the design choices for PGT. Token size: dimensions for group tokens in three grouping blocks. Token shape: number of tokens for group tokens in three grouping blocks. Accuracy measured on ImageNet-1K under linear probe protocal. Results indicate progressively large group token dimensions with flat or descend number of tokens arrangements work the best.}
    \label{tab:design}
  \end{center}
  \vspace{-8mm}
\end{table}

\begin{figure}[h!]
    \centering
    \includegraphics[width=1.0\linewidth]{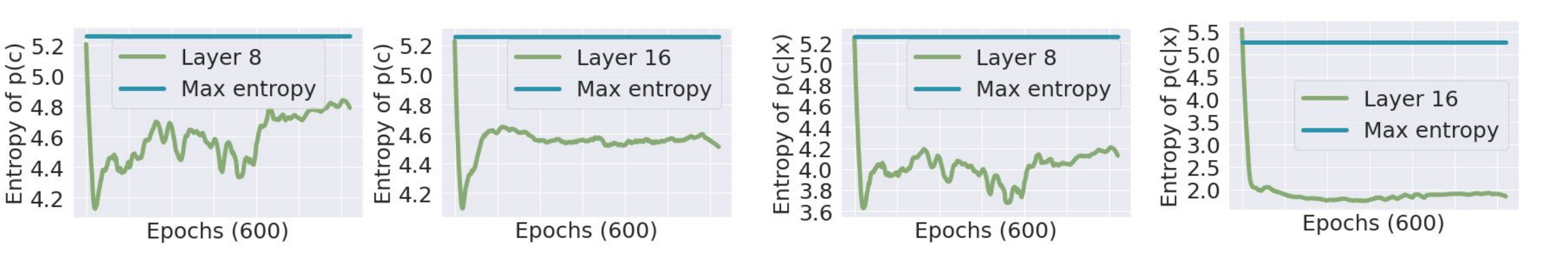}
    \vspace{-6mm}
    \caption{The entropy curves of grouping distributions $p(\bc)$ and $p(\bc|\bx)$ across different layers.}
    \label{fig:curves_grouping_ent}
    \vspace{-2mm}
\end{figure}

\textbf{Grouping distribution entropy.} Will grouping process collapse to some specific group token during training? We visualize the entropy of marginal distribution over tokens $p(\bc)$ and conditional distribution $p(\bc|\bx)$ in figure \ref{fig:curves_grouping_ent}. Interestingly, we observe that conditional probability, i.e. the assignment to group tokens, tends to become more certain during training, while the marginal distribution remains having descend entropy, indicating collapses not happening in training.

\textbf{Peak memory usage.} As discusssed in section \ref{sec:discuss}, given the same number of tokens, the grouping operation uses less memory than the self-attention operation. We show the percentage of peak memory usage in PGT$_{\textsc{G}}$-B compared to ViT-B with the same patch size (4$\times$4) in table \ref{tab:mem}. The usage is obtained from the forward inference graph, as in practice the underlying complex hardware optimizer is a less accurate measurement and varies across infrastructures. 

\begin{table}[h!]
  \begin{center}
    \begin{tabular}{cccccccccc|c}
    \toprule
      \#group tokens & 16 & 32 & 64 & 128 & 256 & 384 & 512 & 768 & 1024 & ViT-B\\
      \midrule
      Peak memory(\%) & 4.6 & 4.6 & 4.6 & 4.6 & 4.6 & 6.1 & 8.2 & 12.2 & 16.3 & 100 \\
      \bottomrule
    \end{tabular}
    \caption{Peak memory usage of PGT-B compared to the baseline model ViT-B with $4\times 4$ patch size.}
    \label{tab:mem}
  \end{center}
  \vspace{-3mm}
\end{table}
\vspace{-4mm}

\subsection{Out-of-distribution adaptive computation}
\label{sec:adapt}
One surprising and powerful ability of PGT is adaptive computation. For example, given a model trained using $M_1$ group tokens per layer, one can choose to use $M_2$ group tokens in inference, where $M_2\neq M_1$. This is because that the initial seeding group tokens are drawn from a probabilistic distribution, and the number of samples can be customized. This property leads to a highly customizable inference without re-training the model. When $M_1\neq M_2$, the model copes with an out-of-distribution (OOD) problem where test time setting is different from training. We observe surprisingly strong generalization with our model. \textit{Specifically, with more tokens $M_2>M_1$ in inference, the performance can actually outperform the setting ($M_2=M_1$) used in training, even if it is OOD for the model.} 

The results for OOD adaptive computation are summarized in table \ref{tab:ood}. We mainly test PGT$_{\textsc{G}}$-Tiny with a grid evaluation that varies the number of group tokens in training $M$ and the number of group tokens in inference $N$, and also show the main model's results in the last row. When using the main model PGT$_{\textsc{G}}$-B to perform adaptive inference, with only 12.5\% of the number of group tokens compared to training, the performance can still be maintained at 72.1\% with only a $\sim$8\% drop on top-1 accuracy. The adaptive computation ability is important for both general image understanding where images have varying number of objects and need different numbers of groups, and scenarios where test-time computational resource is constrained. This flexibility is an important advantage that grouping backbones hold.

\begin{table}[h!]
  \begin{center}
    \begin{tabular}{ccccccc}
    \toprule
      tr/inf & 16 & 32 & 64 & 128 & 256 & 384\\
      \midrule
      PGT$_{\textsc{G}}$-Ti-16 & \underline{57.4} ({\scriptsize$\times 1$ }) & 58.3 ({\scriptsize$\times 2$}) & \textbf{58.5} ({\scriptsize$\times 4$}) & 58.5 ({\scriptsize$\times 8$}) & 58.5({\scriptsize$\times 16$ }) & 58.4 ({\scriptsize$\times 24$}) \\
      PGT$_{\textsc{G}}$-Ti-32 & 57.3 ({\scriptsize$\times \frac{1}{2}$}) & \underline{59.9} ({\scriptsize$\times 1$}) & 60.8 ({\scriptsize$\times 2$}) &  \textbf{61.0} ({\scriptsize$\times 4$}) & 61.0 ({\scriptsize$\times 8$}) &  60.9 ({\scriptsize$\times 12$}) \\
      PGT$_{\textsc{G}}$-Ti-64 & 53.0 ({\scriptsize$\times \frac{1}{4}$}) & 59.2 ({\scriptsize$\times \frac{1}{2}$}) & \underline{61.7} ({\scriptsize$\times 1$}) & 62.6 ({\scriptsize$\times 2$}) & 62.9 ({\scriptsize$\times 4$}) & \textbf{62.9} ({\scriptsize$\times 6$}) \\
      PGT$_{\textsc{G}}$-Ti-128 & 44.9 ({\scriptsize$\times \frac{1}{8}$}) & 56.6 ({\scriptsize$\times \frac{1}{4}$}) & 61.8 ({\scriptsize$\times \frac{1}{2}$}) & \underline{63.9} ({\scriptsize$\times 1$}) & 64.7 ({\scriptsize$\times 2$}) & \textbf{64.8} ({\scriptsize$\times 3$}) \\
      PGT$_{\textsc{G}}$-Ti-256 & 27.2 ({\scriptsize$\times \frac{1}{16}$}) & 47.4 ({\scriptsize$\times \frac{1}{8}$}) & 58.8 ({\scriptsize$\times \frac{1}{4}$}) & 63.3 ({\scriptsize$\times \frac{1}{2}$}) & \underline{65.1} ({\scriptsize$\times 1$}) & \textbf{65.5} ({\scriptsize$\times \frac{3}{2}$}) \\
      PGT$_{\textsc{G}}$-Ti-384 & 26.1 ({\scriptsize$\times \frac{1}{24}$}) & 43.0 ({\scriptsize$\times \frac{1}{12}$}) & 55.4 ({\scriptsize$\times \frac{1}{6}$}) & 61.7 ({\scriptsize$\times \frac{1}{3}$}) & 64.6 ({\scriptsize$\times \frac{2}{3}$}) & \underline{\textbf{65.5}} ({\scriptsize$\times 1$}) \\
      \hline
      PGT$_{\textsc{G}}$-B-256 & 60.4 ({\scriptsize$\times \frac{1}{16}$}) & 72.1 ({\scriptsize$\times \frac{1}{8}$}) & 77.1 ({\scriptsize$\times \frac{1}{4}$}) &  78.9 ({\scriptsize$\times \frac{1}{2}$}) & \underline{79.7} ({\scriptsize$\times 1$}) & \textbf{79.9} ({\scriptsize$\times \frac{3}{2}$}) \\
      \bottomrule
    \end{tabular}
    \caption{Out-of-distribution adaptive computation by selecting different numbers of initially sampled tokens. Row: number of tokens used for training. Column: number of tokens used for inference. Top-1 accuracy is reported under linear evaluation protocol using ImageNet-1K. The reported performance of first six rows is obtained using a tiny version of PGT, and last row is the main model. Number of group tokens is the same for \underline{underlined numbers} in training and inference. \textbf{Bold numbers} are the best results.}
    \label{tab:ood}
  \end{center}
  \vspace{-3mm}
\end{table}

\vspace{-2mm}
\subsection{Downstream task transfer: semantic segmentation on ADE20k}
\label{sec:seg}
To evaluate the generalizability of pretrained feature produced by PGT, we test the transfer performance of semantic segmentation with ADE20k. Following the standard setup, we finetune our model with the same data augmentation for 128 epoch. The baseline method uses DINO + ViT-B/16 \citep{zheng2021rethinking}. For our model, we add one linear classification layer after the pre-trained PGT$_{\textsc{G}}$-B for fine-tuning. To adapt to more objects and complex scenes in the segmentation datasets, we use 1024 group tokens for inference, benefiting from the adaptive computation ability of our model. We find that our model can obtain 45.1\% on mean IoU while the baseline achieves 44.1\% \citep{bao2021beit}, leading to a 1.0\% improvements. 

\vspace{-4mm}
\subsection{Grouping visualization}
\vspace{-2mm}
\label{sec:vis}
We visualize the attention maps calculated between group tokens and input tokens in figure \ref{sec:grouping_vis}. We find that (1) using multiple grouping heads can capture different information within each head. For example, in layer 0, the first head captures light and color, second head focuses on only spatial locations, and the third head potentially relies on textures; (2) group tokens can capture different semantic parts, for example, in the first image, group tokens separate apple, jar, handle, and background. In the second image, camel, legs, camel hump, and human are separately grouped. Compared to standard ViT in DINO \citep{caron2021emerging} where only a single foreground can be extracted using $\textsc{[CLS]}$ token, our model can flexibly group different parts given an image, leading to a set of tokens that are potentially more meaningful and customizable. Note that the grouping results are still different from human's vision, and sometimes generates parts that seem to be ``fragmented''. This is possibly due to the ``parts-to-whole with data augmentation'' training loss. Human vision, in contrast, is sensitive to moving objects and trained within a 4D space. Nevertheless, we believe with a similar dataset, environment and loss design, our grouping model can potentially produce groupings more coherent and sensitive to boundaries and moving objects.

\begin{figure}[t]
    \centering
    \includegraphics[width=1.0\linewidth]{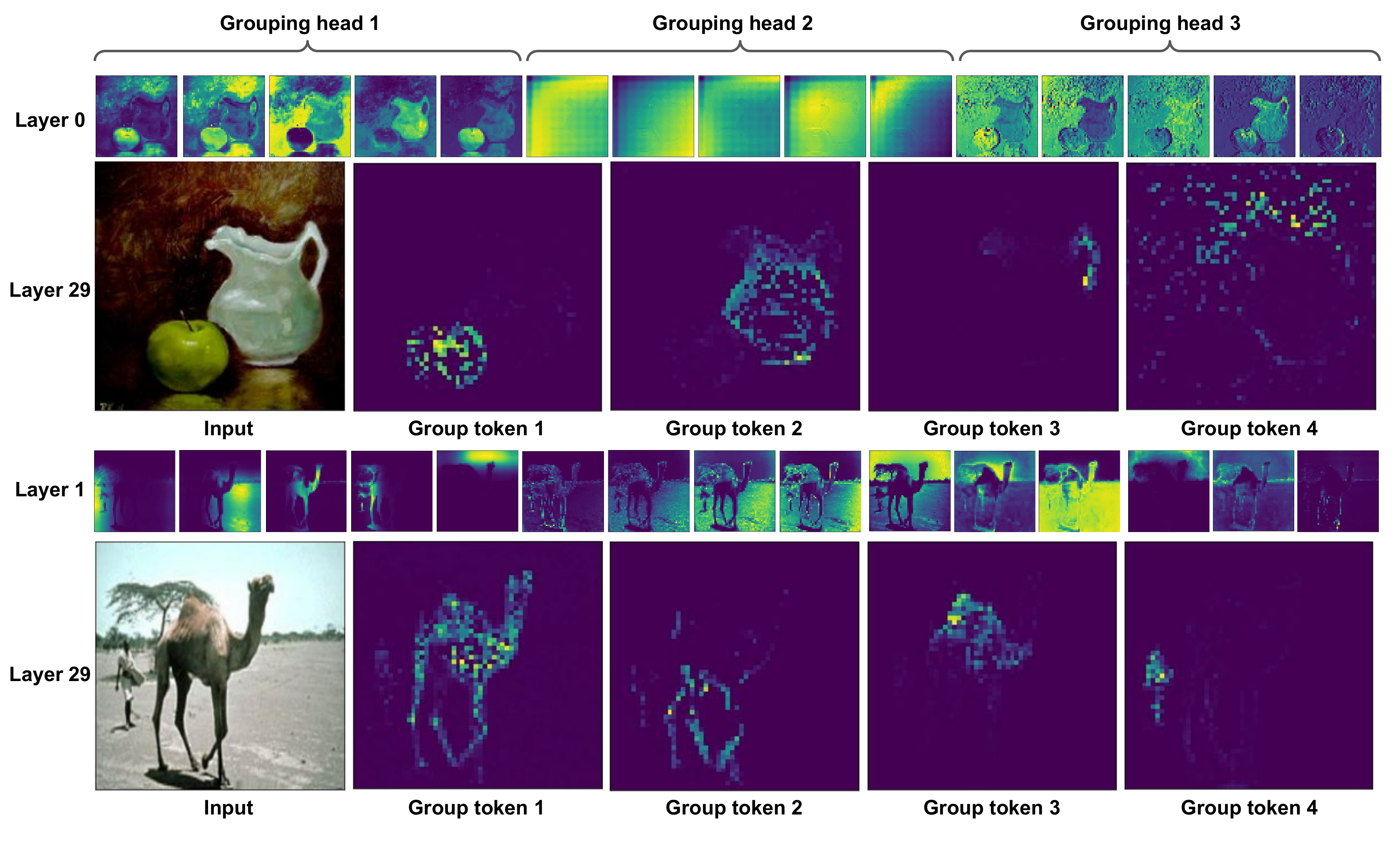}
    \vspace{-6mm}
    \caption{Visualization of attention maps of each group tokens across layers and grouping head. $L$ indicates layer indices. Five group tokens for each grouping head. Smaller images are for early layers, arranged as five group tokens per grouping head. Large images are for the last layer.}
    \label{fig:vis_attn}
    \vspace{-4mm}
\end{figure}

\label{sec:grouping_vis}

\vspace{-3mm}
\section{Conclusion}
\vspace{-3mm}
In this paper, we propose Perceptual Group Tokenizer (PGT), a new visual recognition architecture entirely built through perceptual grouping principles. The proposed model shows strong performance on self-supervised learning benchmark ImageNet-1K with linear probe evaluation, and has desirable properties such as adaptive computation and high model interpretability in each operation. This work can enable a new paradigm for designing visual recognition backbones, and we hope to inspire more research progress along this direction. One limitation of the proposed model is its relatively expensive computation cost due to the iterative grouping processes. This can be potentially addressed by other grouping operations, such as those grouping operations with closed-form solutions, which is a promising direction for the future work.

\bibliography{iclr2024_conference}

\begin{thebibliography}{74}
\providecommand{\natexlab}[1]{#1}
\providecommand{\url}[1]{\texttt{#1}}
\expandafter\ifx\csname urlstyle\endcsname\relax
  \providecommand{\doi}[1]{doi: #1}\else
  \providecommand{\doi}{doi: \begingroup \urlstyle{rm}\Url}\fi

\bibitem[Alemi et~al.(2016)Alemi, Fischer, Dillon, and Murphy]{alemi2016deep}
Alexander~A Alemi, Ian Fischer, Joshua~V Dillon, and Kevin Murphy.
\newblock Deep variational information bottleneck.
\newblock \emph{arXiv preprint arXiv:1612.00410}, 2016.

\bibitem[Arbel{\'a}ez et~al.(2014)Arbel{\'a}ez, Pont-Tuset, Barron, Marques,
  and Malik]{arbelaez2014multiscale}
Pablo Arbel{\'a}ez, Jordi Pont-Tuset, Jonathan~T Barron, Ferran Marques, and
  Jitendra Malik.
\newblock Multiscale combinatorial grouping.
\newblock In \emph{Proceedings of the IEEE conference on computer vision and
  pattern recognition}, pp.\  328--335, 2014.

\bibitem[Bao et~al.(2021)Bao, Dong, Piao, and Wei]{bao2021beit}
Hangbo Bao, Li~Dong, Songhao Piao, and Furu Wei.
\newblock Beit: Bert pre-training of image transformers.
\newblock \emph{arXiv preprint arXiv:2106.08254}, 2021.

\bibitem[Bau et~al.(2017)Bau, Zhou, Khosla, Oliva, and
  Torralba]{bau2017network}
David Bau, Bolei Zhou, Aditya Khosla, Aude Oliva, and Antonio Torralba.
\newblock Network dissection: Quantifying interpretability of deep visual
  representations.
\newblock In \emph{Proceedings of the IEEE conference on computer vision and
  pattern recognition}, pp.\  6541--6549, 2017.

\bibitem[Biza et~al.(2023)Biza, van Steenkiste, Sajjadi, Elsayed, Mahendran,
  and Kipf]{biza2023invariant}
Ondrej Biza, Sjoerd van Steenkiste, Mehdi~SM Sajjadi, Gamaleldin~F Elsayed,
  Aravindh Mahendran, and Thomas Kipf.
\newblock Invariant slot attention: Object discovery with slot-centric
  reference frames.
\newblock \emph{arXiv preprint arXiv:2302.04973}, 2023.

\bibitem[Bolya et~al.(2023)Bolya, Fu, Dai, Zhang, Feichtenhofer, and
  Hoffman]{bolya2023token}
Daniel Bolya, Cheng-Yang Fu, Xiaoliang Dai, Peizhao Zhang, Christoph
  Feichtenhofer, and Judy Hoffman.
\newblock Token merging: Your vit but faster.
\newblock In \emph{The Eleventh International Conference on Learning
  Representations}, 2023.
\newblock URL \url{https://openreview.net/forum?id=JroZRaRw7Eu}.

\bibitem[Burgess et~al.(2019)Burgess, Matthey, Watters, Kabra, Higgins,
  Botvinick, and Lerchner]{burgess2019monet}
Christopher~P Burgess, Loic Matthey, Nicholas Watters, Rishabh Kabra, Irina
  Higgins, Matt Botvinick, and Alexander Lerchner.
\newblock Monet: Unsupervised scene decomposition and representation.
\newblock \emph{arXiv preprint arXiv:1901.11390}, 2019.

\bibitem[Caron et~al.(2020)Caron, Misra, Mairal, Goyal, Bojanowski, and
  Joulin]{caron2020unsupervised}
Mathilde Caron, Ishan Misra, Julien Mairal, Priya Goyal, Piotr Bojanowski, and
  Armand Joulin.
\newblock Unsupervised learning of visual features by contrasting cluster
  assignments.
\newblock \emph{Advances in neural information processing systems},
  33:\penalty0 9912--9924, 2020.

\bibitem[Caron et~al.(2021)Caron, Touvron, Misra, J{\'e}gou, Mairal,
  Bojanowski, and Joulin]{caron2021emerging}
Mathilde Caron, Hugo Touvron, Ishan Misra, Herv{\'e} J{\'e}gou, Julien Mairal,
  Piotr Bojanowski, and Armand Joulin.
\newblock Emerging properties in self-supervised vision transformers.
\newblock In \emph{Proceedings of the IEEE/CVF international conference on
  computer vision}, pp.\  9650--9660, 2021.

\bibitem[Chang et~al.(2022)Chang, Griffiths, and Levine]{chang2022object}
Michael Chang, Tom Griffiths, and Sergey Levine.
\newblock Object representations as fixed points: Training iterative refinement
  algorithms with implicit differentiation.
\newblock \emph{Advances in Neural Information Processing Systems},
  35:\penalty0 32694--32708, 2022.

\bibitem[Chen et~al.(2017)Chen, Papandreou, Kokkinos, Murphy, and
  Yuille]{chen2017deeplab}
Liang-Chieh Chen, George Papandreou, Iasonas Kokkinos, Kevin Murphy, and Alan~L
  Yuille.
\newblock Deeplab: Semantic image segmentation with deep convolutional nets,
  atrous convolution, and fully connected crfs.
\newblock \emph{IEEE transactions on pattern analysis and machine
  intelligence}, 40\penalty0 (4):\penalty0 834--848, 2017.

\bibitem[Chen et~al.(2020{\natexlab{a}})Chen, Kornblith, Norouzi, and
  Hinton]{chen2020simple}
Ting Chen, Simon Kornblith, Mohammad Norouzi, and Geoffrey Hinton.
\newblock A simple framework for contrastive learning of visual
  representations.
\newblock In \emph{International conference on machine learning}, pp.\
  1597--1607. PMLR, 2020{\natexlab{a}}.

\bibitem[Chen et~al.(2020{\natexlab{b}})Chen, Kornblith, Swersky, Norouzi, and
  Hinton]{chen2020big}
Ting Chen, Simon Kornblith, Kevin Swersky, Mohammad Norouzi, and Geoffrey~E
  Hinton.
\newblock Big self-supervised models are strong semi-supervised learners.
\newblock \emph{Advances in neural information processing systems},
  33:\penalty0 22243--22255, 2020{\natexlab{b}}.

\bibitem[Culp et~al.(2022)Culp, Sabour, and Hinton]{culp2022testing}
Laura Culp, Sara Sabour, and Geoffrey~E Hinton.
\newblock Testing glom's ability to infer wholes from ambiguous parts.
\newblock \emph{arXiv preprint arXiv:2211.16564}, 2022.

\bibitem[Dalal \& Triggs(2005)Dalal and Triggs]{dalal2005histograms}
Navneet Dalal and Bill Triggs.
\newblock Histograms of oriented gradients for human detection.
\newblock In \emph{2005 IEEE computer society conference on computer vision and
  pattern recognition (CVPR'05)}, volume~1, pp.\  886--893. Ieee, 2005.

\bibitem[Dao et~al.(2022)Dao, Fu, Ermon, Rudra, and
  R{\'e}]{dao2022flashattention}
Tri Dao, Dan Fu, Stefano Ermon, Atri Rudra, and Christopher R{\'e}.
\newblock Flashattention: Fast and memory-efficient exact attention with
  io-awareness.
\newblock \emph{Advances in Neural Information Processing Systems},
  35:\penalty0 16344--16359, 2022.

\bibitem[Dinh et~al.(2016)Dinh, Sohl-Dickstein, and Bengio]{dinh2016density}
Laurent Dinh, Jascha Sohl-Dickstein, and Samy Bengio.
\newblock Density estimation using real nvp.
\newblock \emph{arXiv preprint arXiv:1605.08803}, 2016.

\bibitem[Dosovitskiy et~al.(2020)Dosovitskiy, Beyer, Kolesnikov, Weissenborn,
  Zhai, Unterthiner, Dehghani, Minderer, Heigold, Gelly,
  et~al.]{dosovitskiy2020image}
Alexey Dosovitskiy, Lucas Beyer, Alexander Kolesnikov, Dirk Weissenborn,
  Xiaohua Zhai, Thomas Unterthiner, Mostafa Dehghani, Matthias Minderer, Georg
  Heigold, Sylvain Gelly, et~al.
\newblock An image is worth 16x16 words: Transformers for image recognition at
  scale.
\newblock \emph{arXiv preprint arXiv:2010.11929}, 2020.

\bibitem[Elsayed et~al.(2022)Elsayed, Mahendran, van Steenkiste, Greff, Mozer,
  and Kipf]{elsayed2022savi++}
Gamaleldin Elsayed, Aravindh Mahendran, Sjoerd van Steenkiste, Klaus Greff,
  Michael~C Mozer, and Thomas Kipf.
\newblock Savi++: Towards end-to-end object-centric learning from real-world
  videos.
\newblock \emph{Advances in Neural Information Processing Systems},
  35:\penalty0 28940--28954, 2022.

\bibitem[Grill et~al.(2020)Grill, Strub, Altch{\'e}, Tallec, Richemond,
  Buchatskaya, Doersch, Avila~Pires, Guo, Gheshlaghi~Azar,
  et~al.]{grill2020bootstrap}
Jean-Bastien Grill, Florian Strub, Florent Altch{\'e}, Corentin Tallec, Pierre
  Richemond, Elena Buchatskaya, Carl Doersch, Bernardo Avila~Pires, Zhaohan
  Guo, Mohammad Gheshlaghi~Azar, et~al.
\newblock Bootstrap your own latent-a new approach to self-supervised learning.
\newblock \emph{Advances in neural information processing systems},
  33:\penalty0 21271--21284, 2020.

\bibitem[He et~al.(2016)He, Zhang, Ren, and Sun]{he2016deep}
Kaiming He, Xiangyu Zhang, Shaoqing Ren, and Jian Sun.
\newblock Deep residual learning for image recognition.
\newblock In \emph{Proceedings of the IEEE conference on computer vision and
  pattern recognition}, pp.\  770--778, 2016.

\bibitem[He et~al.(2022)He, Chen, Xie, Li, Doll{\'a}r, and
  Girshick]{he2022masked}
Kaiming He, Xinlei Chen, Saining Xie, Yanghao Li, Piotr Doll{\'a}r, and Ross
  Girshick.
\newblock Masked autoencoders are scalable vision learners.
\newblock In \emph{Proceedings of the IEEE/CVF conference on computer vision
  and pattern recognition}, pp.\  16000--16009, 2022.

\bibitem[H{\'e}naff et~al.(2022)H{\'e}naff, Koppula, Shelhamer, Zoran, Jaegle,
  Zisserman, Carreira, and Arandjelovi{\'c}]{henaff2022object}
Olivier~J H{\'e}naff, Skanda Koppula, Evan Shelhamer, Daniel Zoran, Andrew
  Jaegle, Andrew Zisserman, Jo{\~a}o Carreira, and Relja Arandjelovi{\'c}.
\newblock Object discovery and representation networks.
\newblock In \emph{European Conference on Computer Vision}, pp.\  123--143.
  Springer, 2022.

\bibitem[Herzog(2018)]{herzog2018perceptual}
Michael~H Herzog.
\newblock Perceptual grouping.
\newblock \emph{Current Biology}, 28\penalty0 (12):\penalty0 R687--R688, 2018.

\bibitem[Hinton(2022)]{hinton2022represent}
Geoffrey Hinton.
\newblock How to represent part-whole hierarchies in a neural network.
\newblock \emph{Neural Computation}, pp.\  1--40, 2022.

\bibitem[Ho et~al.(2020)Ho, Jain, and Abbeel]{ho2020denoising}
Jonathan Ho, Ajay Jain, and Pieter Abbeel.
\newblock Denoising diffusion probabilistic models.
\newblock \emph{Advances in neural information processing systems},
  33:\penalty0 6840--6851, 2020.

\bibitem[Jabri et~al.(2022)Jabri, Fleet, and Chen]{jabri2022scalable}
Allan Jabri, David Fleet, and Ting Chen.
\newblock Scalable adaptive computation for iterative generation.
\newblock \emph{arXiv preprint arXiv:2212.11972}, 2022.

\bibitem[Jaegle et~al.(2021)Jaegle, Borgeaud, Alayrac, Doersch, Ionescu, Ding,
  Koppula, Zoran, Brock, Shelhamer, et~al.]{jaegle2021perceiver}
Andrew Jaegle, Sebastian Borgeaud, Jean-Baptiste Alayrac, Carl Doersch, Catalin
  Ionescu, David Ding, Skanda Koppula, Daniel Zoran, Andrew Brock, Evan
  Shelhamer, et~al.
\newblock Perceiver io: A general architecture for structured inputs \&
  outputs.
\newblock \emph{arXiv preprint arXiv:2107.14795}, 2021.

\bibitem[Ke \& Yu(2022)Ke and Yu]{ke2022cast}
Tsung-Wei Ke and Stella~X Yu.
\newblock Cast: Concurrent recognition and segmentation with adaptive segment
  tokens.
\newblock \emph{arXiv preprint arXiv:2210.00314}, 2022.

\bibitem[Kim et~al.(2019)Kim, Linsley, Thakkar, and
  Serre]{kim2019disentangling}
Junkyung Kim, Drew Linsley, Kalpit Thakkar, and Thomas Serre.
\newblock Disentangling neural mechanisms for perceptual grouping.
\newblock \emph{arXiv preprint arXiv:1906.01558}, 2019.

\bibitem[Kingma \& Welling(2013)Kingma and Welling]{kingma2013auto}
Diederik~P Kingma and Max Welling.
\newblock Auto-encoding variational bayes.
\newblock \emph{arXiv preprint arXiv:1312.6114}, 2013.

\bibitem[Kingma \& Welling(2022)Kingma and Welling]{kingma2022autoencoding}
Diederik~P Kingma and Max Welling.
\newblock Auto-encoding variational bayes, 2022.

\bibitem[Kingma \& Dhariwal(2018)Kingma and Dhariwal]{kingma2018glow}
Durk~P Kingma and Prafulla Dhariwal.
\newblock Glow: Generative flow with invertible 1x1 convolutions.
\newblock \emph{Advances in neural information processing systems}, 31, 2018.

\bibitem[Kr{\"a}henb{\"u}hl \& Koltun(2011)Kr{\"a}henb{\"u}hl and
  Koltun]{krahenbuhl2011efficient}
Philipp Kr{\"a}henb{\"u}hl and Vladlen Koltun.
\newblock Efficient inference in fully connected crfs with gaussian edge
  potentials.
\newblock \emph{Advances in neural information processing systems}, 24, 2011.

\bibitem[Krizhevsky et~al.(2012)Krizhevsky, Sutskever, and
  Hinton]{krizhevsky2012imagenet}
Alex Krizhevsky, Ilya Sutskever, and Geoffrey~E Hinton.
\newblock Imagenet classification with deep convolutional neural networks.
\newblock \emph{Advances in neural information processing systems}, 25, 2012.

\bibitem[LeCun et~al.(1998)LeCun, Bottou, Bengio, and
  Haffner]{lecun1998gradient}
Yann LeCun, L{\'e}on Bottou, Yoshua Bengio, and Patrick Haffner.
\newblock Gradient-based learning applied to document recognition.
\newblock \emph{Proceedings of the IEEE}, 86\penalty0 (11):\penalty0
  2278--2324, 1998.

\bibitem[Levinshtein et~al.(2013)Levinshtein, Sminchisescu, and
  Dickinson]{levinshtein2013multiscale}
Alex Levinshtein, Cristian Sminchisescu, and Sven Dickinson.
\newblock Multiscale symmetric part detection and grouping.
\newblock \emph{International journal of computer vision}, 104:\penalty0
  117--134, 2013.

\bibitem[Liu et~al.(2022{\natexlab{a}})Liu, Wu, Liu, and Guo]{liu2022dynamic}
Kai Liu, Tianyi Wu, Cong Liu, and Guodong Guo.
\newblock Dynamic group transformer: A general vision transformer backbone with
  dynamic group attention.
\newblock \emph{arXiv preprint arXiv:2203.03937}, 2022{\natexlab{a}}.

\bibitem[Liu et~al.(2021)Liu, Lin, Cao, Hu, Wei, Zhang, Lin, and
  Guo]{liu2021swin}
Ze~Liu, Yutong Lin, Yue Cao, Han Hu, Yixuan Wei, Zheng Zhang, Stephen Lin, and
  Baining Guo.
\newblock Swin transformer: Hierarchical vision transformer using shifted
  windows.
\newblock In \emph{Proceedings of the IEEE/CVF international conference on
  computer vision}, pp.\  10012--10022, 2021.

\bibitem[Liu et~al.(2022{\natexlab{b}})Liu, Mao, Wu, Feichtenhofer, Darrell,
  and Xie]{liu2022convnet}
Zhuang Liu, Hanzi Mao, Chao-Yuan Wu, Christoph Feichtenhofer, Trevor Darrell,
  and Saining Xie.
\newblock A convnet for the 2020s.
\newblock In \emph{Proceedings of the IEEE/CVF conference on computer vision
  and pattern recognition}, pp.\  11976--11986, 2022{\natexlab{b}}.

\bibitem[Locatello et~al.(2020)Locatello, Weissenborn, Unterthiner, Mahendran,
  Heigold, Uszkoreit, Dosovitskiy, and Kipf]{locatello2020object}
Francesco Locatello, Dirk Weissenborn, Thomas Unterthiner, Aravindh Mahendran,
  Georg Heigold, Jakob Uszkoreit, Alexey Dosovitskiy, and Thomas Kipf.
\newblock Object-centric learning with slot attention.
\newblock \emph{Advances in Neural Information Processing Systems},
  33:\penalty0 11525--11538, 2020.

\bibitem[Loshchilov \& Hutter(2018)Loshchilov and Hutter]{loshchilov2018fixing}
Ilya Loshchilov and Frank Hutter.
\newblock Fixing weight decay regularization in adam.
\newblock 2018.

\bibitem[Lowe(2004)]{lowe2004distinctive}
David~G Lowe.
\newblock Distinctive image features from scale-invariant keypoints.
\newblock \emph{International journal of computer vision}, 60:\penalty0
  91--110, 2004.

\bibitem[Ma et~al.(2023)Ma, Zhou, Wang, Qin, Sun, Liu, and Fu]{ma2023image}
Xu~Ma, Yuqian Zhou, Huan Wang, Can Qin, Bin Sun, Chang Liu, and Yun Fu.
\newblock Image as set of points.
\newblock In \emph{The Eleventh International Conference on Learning
  Representations}, 2023.
\newblock URL \url{https://openreview.net/forum?id=awnvqZja69}.

\bibitem[Ma et~al.(2007)Ma, Derksen, Hong, and Wright]{ma2007segmentation}
Yi~Ma, Harm Derksen, Wei Hong, and John Wright.
\newblock Segmentation of multivariate mixed data via lossy data coding and
  compression.
\newblock \emph{IEEE transactions on pattern analysis and machine
  intelligence}, 29\penalty0 (9):\penalty0 1546--1562, 2007.

\bibitem[Marino et~al.(2018)Marino, Yue, and Mandt]{marino2018iterative}
Joe Marino, Yisong Yue, and Stephan Mandt.
\newblock Iterative amortized inference.
\newblock In \emph{International Conference on Machine Learning}, pp.\
  3403--3412. PMLR, 2018.

\bibitem[Oord et~al.(2018)Oord, Li, and Vinyals]{oord2018representation}
Aaron van~den Oord, Yazhe Li, and Oriol Vinyals.
\newblock Representation learning with contrastive predictive coding.
\newblock \emph{arXiv preprint arXiv:1807.03748}, 2018.

\bibitem[Oquab et~al.(2023)Oquab, Darcet, Moutakanni, Vo, Szafraniec, Khalidov,
  Fernandez, Haziza, Massa, El-Nouby, et~al.]{oquab2023dinov2}
Maxime Oquab, Timoth{\'e}e Darcet, Th{\'e}o Moutakanni, Huy Vo, Marc
  Szafraniec, Vasil Khalidov, Pierre Fernandez, Daniel Haziza, Francisco Massa,
  Alaaeldin El-Nouby, et~al.
\newblock Dinov2: Learning robust visual features without supervision.
\newblock \emph{arXiv preprint arXiv:2304.07193}, 2023.

\bibitem[Palmer(2002)]{palmer2002perceptual}
Stephen~E Palmer.
\newblock Perceptual grouping: It's later than you think.
\newblock \emph{Current Directions in Psychological Science}, 11\penalty0
  (3):\penalty0 101--106, 2002.

\bibitem[Peng et~al.(2022)Peng, Dong, Bao, Ye, and Wei]{peng2022beit}
Zhiliang Peng, Li~Dong, Hangbo Bao, Qixiang Ye, and Furu Wei.
\newblock Beit v2: Masked image modeling with vector-quantized visual
  tokenizers.
\newblock \emph{arXiv preprint arXiv:2208.06366}, 2022.

\bibitem[Pont-Tuset et~al.(2016)Pont-Tuset, Arbelaez, Barron, Marques, and
  Malik]{pont2016multiscale}
Jordi Pont-Tuset, Pablo Arbelaez, Jonathan~T Barron, Ferran Marques, and
  Jitendra Malik.
\newblock Multiscale combinatorial grouping for image segmentation and object
  proposal generation.
\newblock \emph{IEEE transactions on pattern analysis and machine
  intelligence}, 39\penalty0 (1):\penalty0 128--140, 2016.

\bibitem[Qi et~al.(2020)Qi, You, Wang, Ma, and Malik]{qi2020deep}
Haozhi Qi, Chong You, Xiaolong Wang, Yi~Ma, and Jitendra Malik.
\newblock Deep isometric learning for visual recognition.
\newblock In \emph{International conference on machine learning}, pp.\
  7824--7835. PMLR, 2020.

\bibitem[Rao et~al.(2021)Rao, Zhao, Liu, Lu, Zhou, and
  Hsieh]{rao2021dynamicvit}
Yongming Rao, Wenliang Zhao, Benlin Liu, Jiwen Lu, Jie Zhou, and Cho-Jui Hsieh.
\newblock Dynamicvit: Efficient vision transformers with dynamic token
  sparsification.
\newblock \emph{Advances in neural information processing systems},
  34:\penalty0 13937--13949, 2021.

\bibitem[Reddy et~al.(2021)Reddy, Dragan, and Levine]{reddy2021pragmatic}
Sid Reddy, Anca Dragan, and Sergey Levine.
\newblock Pragmatic image compression for human-in-the-loop decision-making.
\newblock \emph{Advances in Neural Information Processing Systems},
  34:\penalty0 26499--26510, 2021.

\bibitem[Rosten et~al.(2008)Rosten, Porter, and Drummond]{rosten2008faster}
Edward Rosten, Reid Porter, and Tom Drummond.
\newblock Faster and better: A machine learning approach to corner detection.
\newblock \emph{IEEE transactions on pattern analysis and machine
  intelligence}, 32\penalty0 (1):\penalty0 105--119, 2008.

\bibitem[Russakovsky et~al.(2015)Russakovsky, Deng, Su, Krause, Satheesh, Ma,
  Huang, Karpathy, Khosla, Bernstein, et~al.]{russakovsky2015imagenet}
Olga Russakovsky, Jia Deng, Hao Su, Jonathan Krause, Sanjeev Satheesh, Sean Ma,
  Zhiheng Huang, Andrej Karpathy, Aditya Khosla, Michael Bernstein, et~al.
\newblock Imagenet large scale visual recognition challenge.
\newblock \emph{International journal of computer vision}, 115:\penalty0
  211--252, 2015.

\bibitem[Seitzer et~al.(2022)Seitzer, Horn, Zadaianchuk, Zietlow, Xiao,
  Simon-Gabriel, He, Zhang, Sch{\"o}lkopf, Brox, et~al.]{seitzer2022bridging}
Maximilian Seitzer, Max Horn, Andrii Zadaianchuk, Dominik Zietlow, Tianjun
  Xiao, Carl-Johann Simon-Gabriel, Tong He, Zheng Zhang, Bernhard
  Sch{\"o}lkopf, Thomas Brox, et~al.
\newblock Bridging the gap to real-world object-centric learning.
\newblock In \emph{The Eleventh International Conference on Learning
  Representations}, 2022.

\bibitem[Shi \& Malik(2000)Shi and Malik]{shi2000normalized}
Jianbo Shi and Jitendra Malik.
\newblock Normalized cuts and image segmentation.
\newblock \emph{IEEE Transactions on pattern analysis and machine
  intelligence}, 22\penalty0 (8):\penalty0 888--905, 2000.

\bibitem[Simonyan \& Zisserman(2014)Simonyan and Zisserman]{simonyan2014very}
Karen Simonyan and Andrew Zisserman.
\newblock Very deep convolutional networks for large-scale image recognition.
\newblock \emph{arXiv preprint arXiv:1409.1556}, 2014.

\bibitem[Szegedy et~al.(2015)Szegedy, Liu, Jia, Sermanet, Reed, Anguelov,
  Erhan, Vanhoucke, and Rabinovich]{szegedy2015going}
Christian Szegedy, Wei Liu, Yangqing Jia, Pierre Sermanet, Scott Reed, Dragomir
  Anguelov, Dumitru Erhan, Vincent Vanhoucke, and Andrew Rabinovich.
\newblock Going deeper with convolutions.
\newblock In \emph{Proceedings of the IEEE conference on computer vision and
  pattern recognition}, pp.\  1--9, 2015.

\bibitem[Tan \& Le(2019)Tan and Le]{tan2019efficientnet}
Mingxing Tan and Quoc Le.
\newblock Efficientnet: Rethinking model scaling for convolutional neural
  networks.
\newblock In \emph{International conference on machine learning}, pp.\
  6105--6114. PMLR, 2019.

\bibitem[Tian et~al.(2020)Tian, Krishnan, and Isola]{tian2020contrastive}
Yonglong Tian, Dilip Krishnan, and Phillip Isola.
\newblock Contrastive multiview coding.
\newblock In \emph{Computer Vision--ECCV 2020: 16th European Conference,
  Glasgow, UK, August 23--28, 2020, Proceedings, Part XI 16}, pp.\  776--794.
  Springer, 2020.

\bibitem[Touvron et~al.(2021)Touvron, Cord, Douze, Massa, Sablayrolles, and
  J{\'e}gou]{touvron2021training}
Hugo Touvron, Matthieu Cord, Matthijs Douze, Francisco Massa, Alexandre
  Sablayrolles, and Herv{\'e} J{\'e}gou.
\newblock Training data-efficient image transformers \& distillation through
  attention.
\newblock In \emph{International conference on machine learning}, pp.\
  10347--10357. PMLR, 2021.

\bibitem[Uijlings et~al.(2013)Uijlings, Van De~Sande, Gevers, and
  Smeulders]{uijlings2013selective}
Jasper~RR Uijlings, Koen~EA Van De~Sande, Theo Gevers, and Arnold~WM Smeulders.
\newblock Selective search for object recognition.
\newblock \emph{International journal of computer vision}, 104:\penalty0
  154--171, 2013.

\bibitem[Vaswani et~al.(2017)Vaswani, Shazeer, Parmar, Uszkoreit, Jones, Gomez,
  Kaiser, and Polosukhin]{vaswani2017attention}
Ashish Vaswani, Noam Shazeer, Niki Parmar, Jakob Uszkoreit, Llion Jones,
  Aidan~N Gomez, {\L}ukasz Kaiser, and Illia Polosukhin.
\newblock Attention is all you need.
\newblock \emph{Advances in neural information processing systems}, 30, 2017.

\bibitem[Wagemans et~al.(2012)Wagemans, Elder, Kubovy, Palmer, Peterson, Singh,
  and Von~der Heydt]{wagemans2012century}
Johan Wagemans, James~H Elder, Michael Kubovy, Stephen~E Palmer, Mary~A
  Peterson, Manish Singh, and R{\"u}diger Von~der Heydt.
\newblock A century of gestalt psychology in visual perception: I. perceptual
  grouping and figure--ground organization.
\newblock \emph{Psychological bulletin}, 138\penalty0 (6):\penalty0 1172, 2012.

\bibitem[Wu et~al.(2022)Wu, Dvornik, Greff, Kipf, and Garg]{wu2022slotformer}
Ziyi Wu, Nikita Dvornik, Klaus Greff, Thomas Kipf, and Animesh Garg.
\newblock Slotformer: Unsupervised visual dynamics simulation with
  object-centric models.
\newblock In \emph{The Eleventh International Conference on Learning
  Representations}, 2022.

\bibitem[Xu et~al.(2022)Xu, De~Mello, Liu, Byeon, Breuel, Kautz, and
  Wang]{xu2022groupvit}
Jiarui Xu, Shalini De~Mello, Sifei Liu, Wonmin Byeon, Thomas Breuel, Jan Kautz,
  and Xiaolong Wang.
\newblock Groupvit: Semantic segmentation emerges from text supervision.
\newblock In \emph{Proceedings of the IEEE/CVF Conference on Computer Vision
  and Pattern Recognition}, pp.\  18134--18144, 2022.

\bibitem[Yin et~al.(2022)Yin, Vahdat, Alvarez, Mallya, Kautz, and
  Molchanov]{yin2022vit}
Hongxu Yin, Arash Vahdat, Jose~M Alvarez, Arun Mallya, Jan Kautz, and Pavlo
  Molchanov.
\newblock A-vit: Adaptive tokens for efficient vision transformer.
\newblock In \emph{Proceedings of the IEEE/CVF Conference on Computer Vision
  and Pattern Recognition}, pp.\  10809--10818, 2022.

\bibitem[Yosinski et~al.(2015)Yosinski, Clune, Nguyen, Fuchs, and
  Lipson]{yosinski2015understanding}
Jason Yosinski, Jeff Clune, Anh Nguyen, Thomas Fuchs, and Hod Lipson.
\newblock Understanding neural networks through deep visualization.
\newblock \emph{arXiv preprint arXiv:1506.06579}, 2015.

\bibitem[Zeiler \& Fergus(2014)Zeiler and Fergus]{zeiler2014visualizing}
Matthew~D Zeiler and Rob Fergus.
\newblock Visualizing and understanding convolutional networks.
\newblock In \emph{Computer Vision--ECCV 2014: 13th European Conference,
  Zurich, Switzerland, September 6-12, 2014, Proceedings, Part I 13}, pp.\
  818--833. Springer, 2014.

\bibitem[Zheng et~al.(2021)Zheng, Lu, Zhao, Zhu, Luo, Wang, Fu, Feng, Xiang,
  Torr, et~al.]{zheng2021rethinking}
Sixiao Zheng, Jiachen Lu, Hengshuang Zhao, Xiatian Zhu, Zekun Luo, Yabiao Wang,
  Yanwei Fu, Jianfeng Feng, Tao Xiang, Philip~HS Torr, et~al.
\newblock Rethinking semantic segmentation from a sequence-to-sequence
  perspective with transformers.
\newblock In \emph{Proceedings of the IEEE/CVF conference on computer vision
  and pattern recognition}, pp.\  6881--6890, 2021.

\bibitem[Zhou et~al.(2014)Zhou, Khosla, Lapedriza, Oliva, and
  Torralba]{zhou2014object}
Bolei Zhou, Aditya Khosla, Agata Lapedriza, Aude Oliva, and Antonio Torralba.
\newblock Object detectors emerge in deep scene cnns.
\newblock \emph{arXiv preprint arXiv:1412.6856}, 2014.

\bibitem[Zhou et~al.(2021)Zhou, Wei, Wang, Shen, Xie, Yuille, and
  Kong]{zhou2021ibot}
Jinghao Zhou, Chen Wei, Huiyu Wang, Wei Shen, Cihang Xie, Alan Yuille, and Tao
  Kong.
\newblock ibot: Image bert pre-training with online tokenizer.
\newblock \emph{arXiv preprint arXiv:2111.07832}, 2021.

\end{thebibliography}
\bibliographystyle{iclr2024_conference}

\newpage 

\appendix
\section{Appendix}

\subsection{Learnable sampling distributions}
\label{sec:init_dist}
Our proposed Perceptual Group Tokenizer (PGT) model initializes a set of group tokens through sampling from a distribution. This set of group tokens then serve as the initial ``seeding'' for the grouping process. We explore two methods to serve as the initial distribution: learnable Gaussian distribution and Normalizing Flows. We would like the extra cost of the grouping process to be minimal, therefore, use two light-weighted versions.
\subsubsection{Gaussian}
Similar to the standard usage of learnable Gaussians in generative model literature \citep{kingma2022autoencoding, ho2020denoising}, we use the reparameterization to perform a learnable sampling process: $\bc = \bmu + \bsigma * \bepsilon$, where $\bepsilon$ is drawn from a unit Gaussian $\mathcal{N}(0, I)$
\subsubsection{Flow}
As Gaussian distribution might have limitations in covering complex distribution shapes, especially in the high-dimentional space, we also explore a version with one step of affine coupling flow transformation \citep{dinh2016density}. Since we only require the differentiable sampling procedure and do not need to compute the determinant of Jacobian matrix, we directly apply the transformation without splitting the dimensions by half:
\begin{eqnarray}
\label{eqn:flow}
    \bc &=& \ba * \bepsilon + \bb \\
    (\log \bs, \bt) &=& \text{MLP}(\bc) \\
    \bs &=& \exp(\log \bs) \\
    \bc &=& \bs * \bc + \bt
\end{eqnarray}
where $\bepsilon$ is drawn from a unit Gaussian $\mathcal{N}(0, I)$. This transformation is simply a re-scaling and translation (similar to Gaussian) but conditioned on per sample $\bepsilon$. More details are in \citep{dinh2016density, kingma2018glow}. We only apply one step of this transformation, leading to minimal parameter increase and negligible inference time difference.

\subsection{Model analysis}
In this section, we add more analysis on our model's performance and computational costs.
\subsubsection{Grouping entropy}

\begin{figure}[h!]
    \centering
    \includegraphics[width=1.0\linewidth]{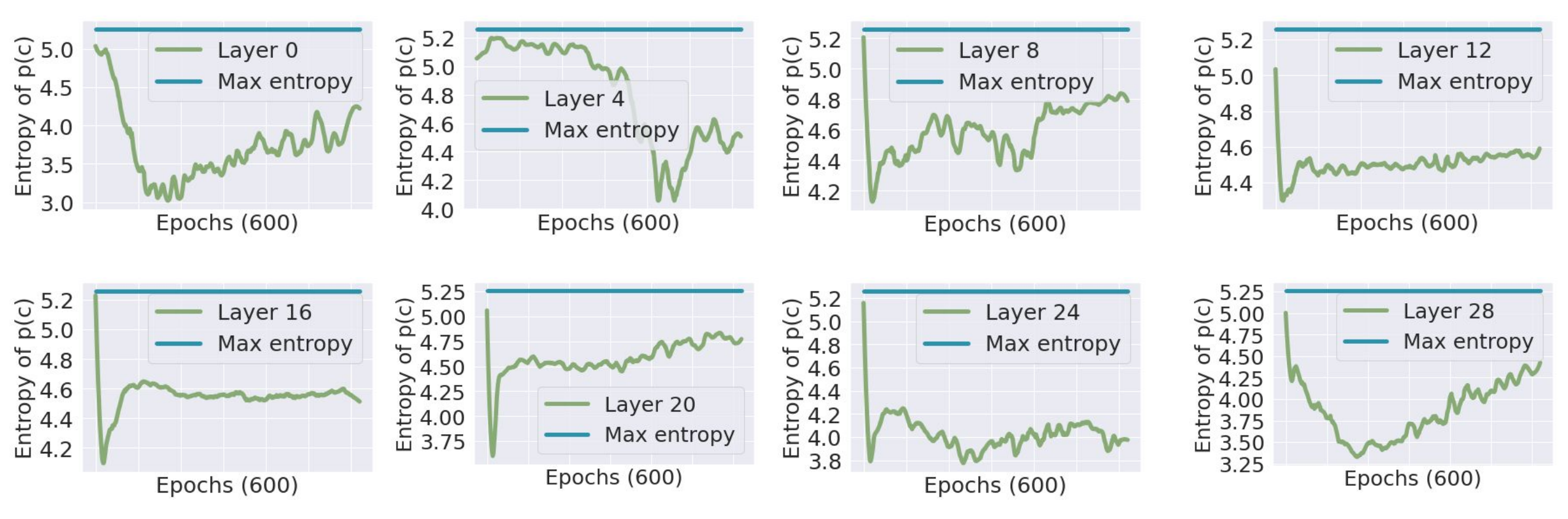}
    \caption{The entropy curves of marginal distribution $p(\bc)$ grouping across different layers.}
    \label{fig:curves_p_ent}
\end{figure}

\begin{figure}[h!]
    \centering
    \includegraphics[width=1.0\linewidth]{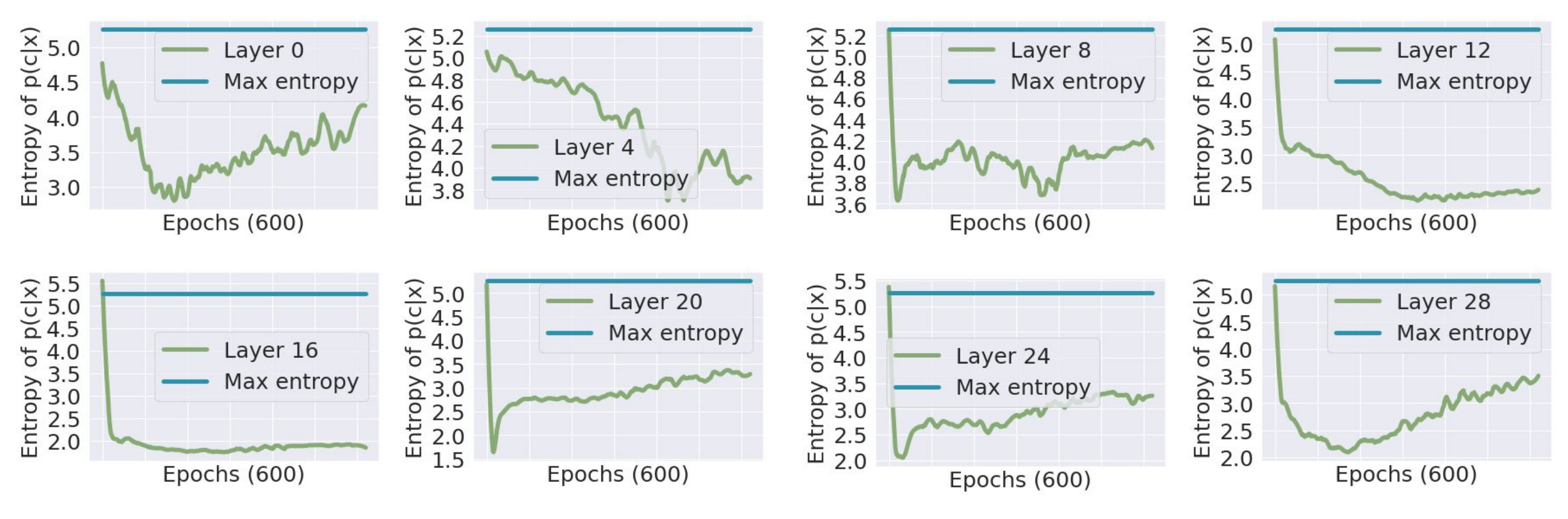}
    \caption{The entropy curves of conditional distribution $p(\bc|\bx)$ grouping across different layers.}
    \label{fig:curves_cond_p_ent}
\end{figure}

\textbf{Grouping distribution entropy.} The main paper has discussed and shown the grouping distribution entropy curves on several layers. In the appendix, we demonstrate curves from more layers in figure \ref{fig:curves_p_ent} and figure \ref{fig:curves_cond_p_ent}, where the first one is marginal distribution and second one is conditional distribution.

\subsubsection{Grouping iterations}

In our backbone, we find that more grouping iterations will lead to a better performance. We explore the number of grouping iterations on the PGT-Tiny model and PGT$_{\textsc{G}}$-B-256. On the tiny version, we find the model achieves 61.4, 63.8, and 65.1 on the linear probe evaluation with number of interations is 1, 2, and 3. For the main model, the performances are 79.3, 79.6, and 79.7 respectively. The model's increased depth potentially helps with the lack of grouping iterations in the deep model. But in general, having the grouping process is still important in obtaining higher performance.

\subsubsection{Inference time}

We also profile our model's inference time, compared with ViT-B with 4x4 patches (the same amount tokens) for ablation study on the grouping operation. Note that our model and framework are built upon a complex infrastructure that uses XLA and other hardware accelerator to optimize speed. We find varying number of group tokens only lead to small influence. PGT-B-256 has ~640 im/sec/core and ViT-B/4 has ~680 im/sec/core. Using smaller number of grouping iterations can speed up the inference to ~710 im/sec/core (2 iter2) and ~820 im/sec/core (1 iter). 

Note that this is only due to the specialty of the underlying infrastructure. In general, having less number of group tokens should still increase the inference speed, since the attention operation is a key computation bottleneck for vision models. 

\subsubsection{Gflops}

In table \ref{tab:gflops}, we show the gflops for our model under various inference budgets. Note that, as pointed in other works \citep{dao2022flashattention}, gflops often do not fully reflect the model's computation performance. Due to that our model needs iterative grouping process, it'll increase the gflops count. But as shown in peak memory usage and inference time, the model's computation costs are either similar are much less.

\begin{table}[h!]
  \begin{center}
    \label{tab:gflops}
    \begin{tabular}{cccccccccc|c}
    \toprule
      \#group tokens & 16 & 32 & 64 & 128 & 256 & 384 & 512 & 768 & 1024 & ViT-B\\
      \midrule
      Gflops & 99.2 & 131.1 & 194.9 & 322.6 & 577.8 & 833.1 & 1088.3 & 1599.0 & 2109.3 & 451.6 \\
      \bottomrule
    \end{tabular}
    \caption{Gflops of PGT-B compared to the baseline model ViT-B with the same patch size.}
  \end{center}
\end{table}

\subsubsection{Probabilistic perspective of grouping operations}
Due to the probabilistic nature, our model is also quite compatible with a full ``treatment'' with the variational inference framework, which can provide certain backup for our grouping operations already in the current model. We can treat the group token embeddings $\bc$ as the latent variables, where the grouping process uses iterative amortized inference \citep{marino2018iterative} to refine the latent variable. The grouping modules, including GRU, MLP, attention, and other layers are designed to better infer the embeddings (latent variables). The training signal is a pragmatic loss (instead of reconstruction loss), which has been demonstrated in \citep{reddy2021pragmatic, alemi2016deep}. The key differences are: (1) there is no sampling in each inference step; (2) the regularization from unit Gaussian distribution is set to zero. We do believe a full probabilistic treatment of the perceptual grouping architecture can be a very interesting next step.

\subsubsection{More visualizations}

\begin{figure}[h!]
    \centering
    \includegraphics[width=1.0\linewidth]{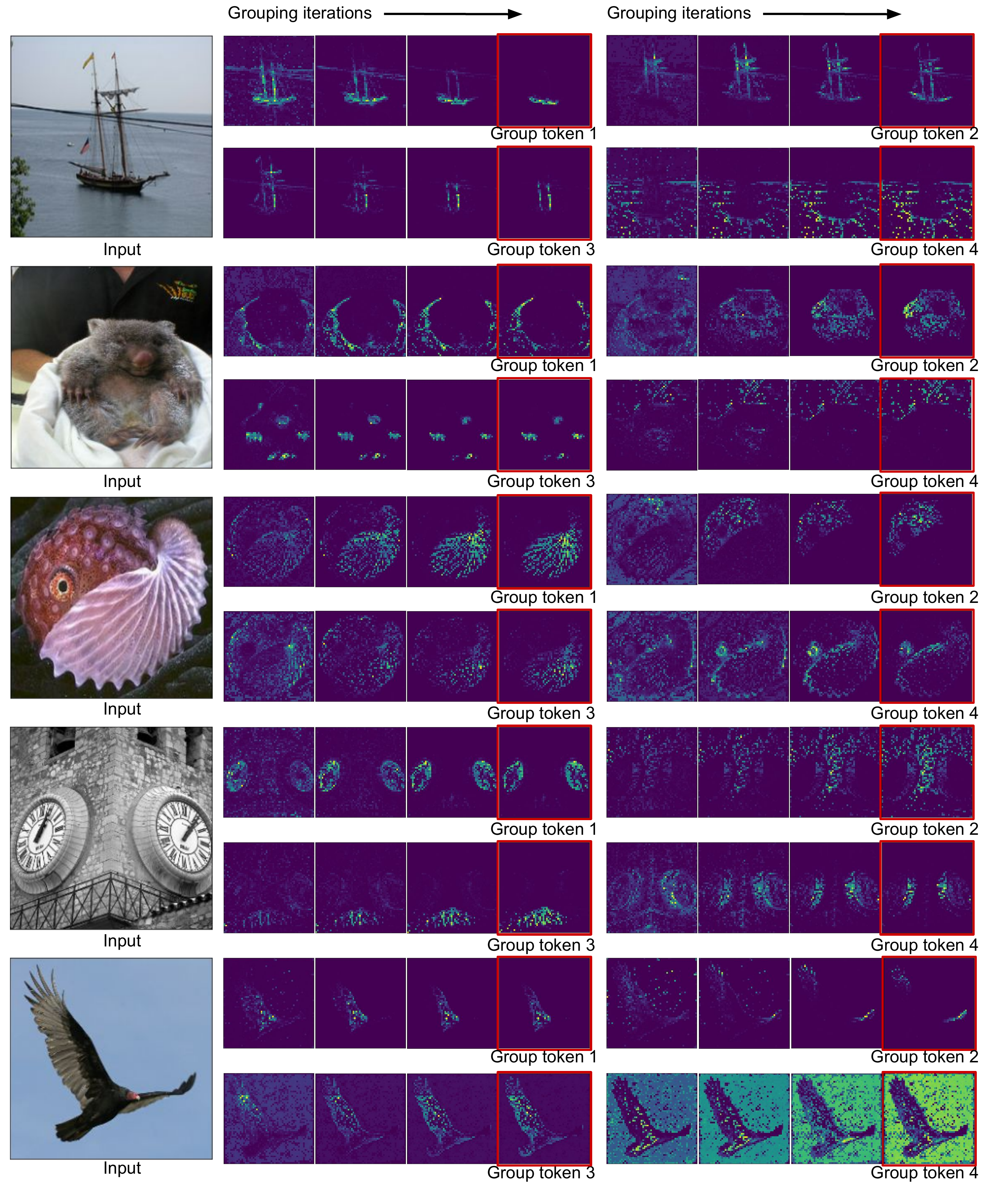}
    \caption{Visualization for attention maps of group token samples during the grouping process. PGT uses 256 group tokens in inference time.}
    \label{fig:vis_iter_more}
    
\end{figure}

\begin{figure}[h!]
    \centering
    \includegraphics[width=1.0\linewidth]{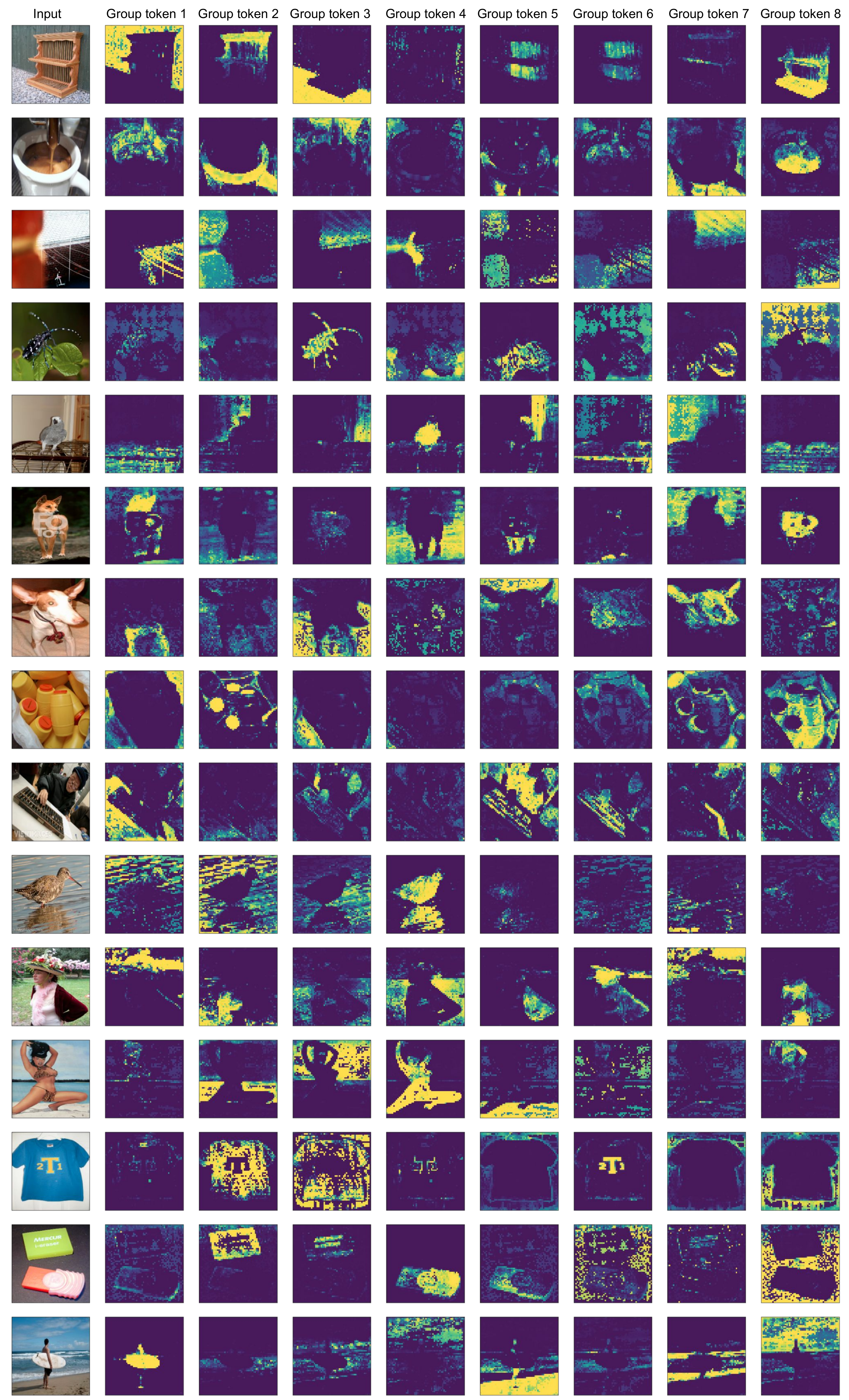}
    \vspace{-8mm}
    \caption{Grouping results from the 21st layer, using 8 group tokens in inference time.}
    \label{fig:vis_group_parts}
\end{figure}

\begin{figure}[h!]
    \centering
    \includegraphics[width=1.0\linewidth]{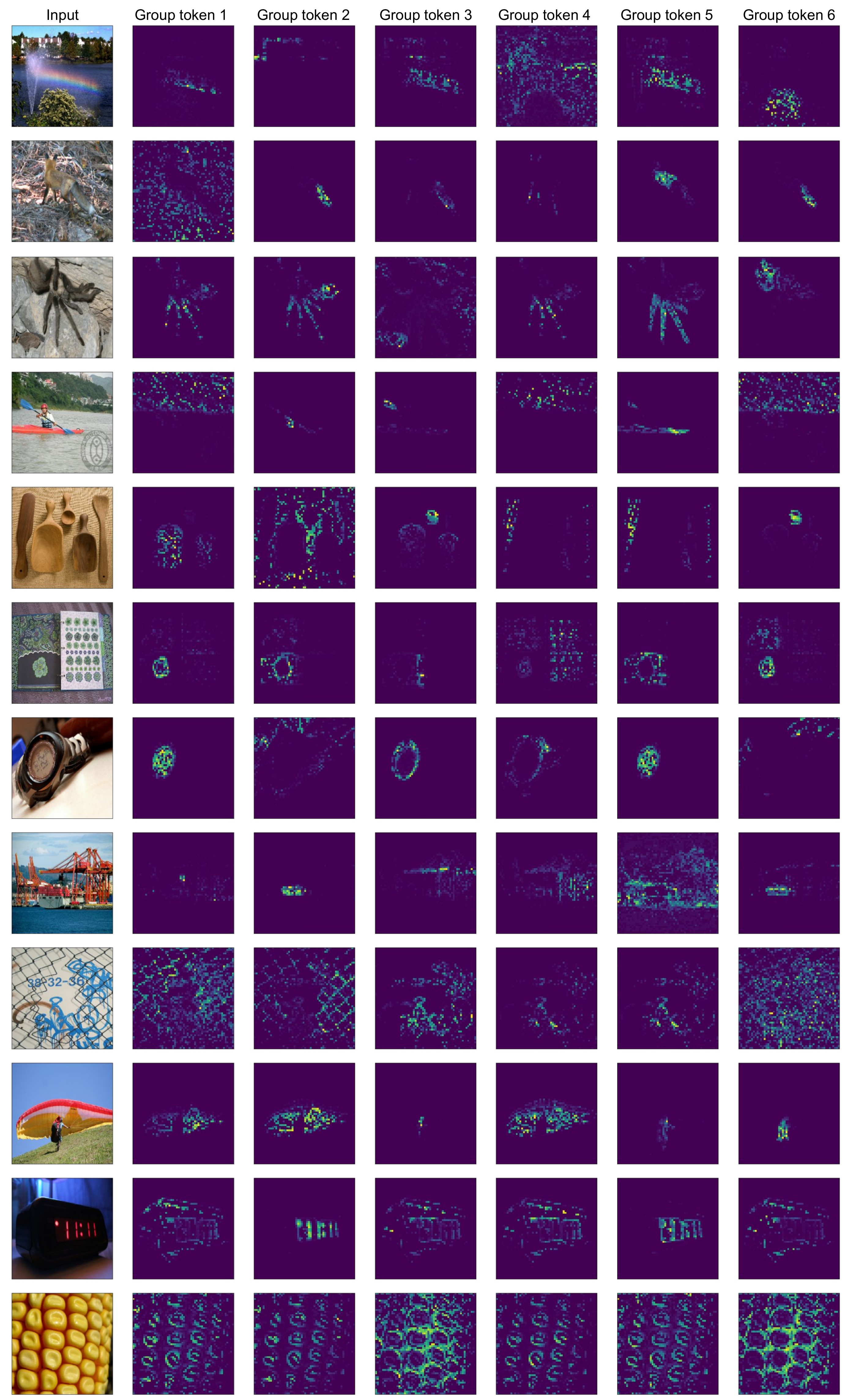}
    \vspace{-8mm}
    \caption{Grouping results from the last layer, using 256 group tokens in inference time.}
    \label{fig:vis_group_parts_8_tokens}
\end{figure}

In this section, we show more visualizations of the attention maps for generated group tokens by Perceptual Group Tokenizers in figure \ref{fig:vis_iter_more}, \ref{fig:vis_group_parts} and \ref{fig:vis_group_parts_8_tokens}.

\end{document}